%% file: neurips_2026.tex
\documentclass{article}

    \PassOptionsToPackage{numbers, compress}{natbib}

 \usepackage[main, final]{neurips_2026}

\usepackage[utf8]{inputenc} 
\usepackage[T1]{fontenc}    
\usepackage{hyperref}       
\usepackage{url}            
\usepackage{booktabs}       
\usepackage{amsfonts}       
\usepackage{nicefrac}       
\usepackage{microtype}      
\usepackage{xcolor}         
\usepackage{amsmath}
\usepackage{graphicx}
\usepackage{pifont}
\usepackage{booktabs}
\usepackage{caption}
\usepackage{algorithm}
\usepackage{algpseudocode}
\usepackage{float}
\usepackage{wrapfig}
\usepackage{subcaption}
\usepackage[table]{xcolor}

\usepackage[most]{tcolorbox}
\usepackage{xcolor}
\usepackage{fvextra}

\usepackage{inconsolata}

\usepackage{booktabs}
\usepackage{multirow}

\definecolor{headergray}{RGB}{135,135,135}
\definecolor{lightpurple}{RGB}{236,236,255}
\definecolor{lightblue}{RGB}{249,253,254}
\definecolor{lightgreen}{RGB}{250,254,252}
\definecolor{lightgray}{RGB}{242,242,242}

\newtcbox{\prompttitlebox}{
  enhanced,
  colback=gray!80,
  colframe=gray!80,
  boxrule=0pt,
  arc=0pt,
  left=6pt,
  right=6pt,
  top=2pt,
  bottom=2pt,
  fontupper=\ttfamily\bfseries\footnotesize,
  colupper=white,
  on line
}

\newtcolorbox{promptblock}[1][white]{
  enhanced jigsaw,
  breakable,
  colback=#1,
  colframe=black!45,
  boxrule=0.4pt,
  arc=0pt,
  left=4pt,
  right=4pt,
  top=3pt,
  bottom=3pt,
  width=\linewidth,
  before=\par\noindent,
  after=\par\smallskip,
  before upper={\parindent0pt}
}

\newcommand{\cmark}{\ding{51}}

\title{AgentGrad: Intervention-guided Prompt Optimization for Multi-Agent Systems}

\author{
  Jaewon Chu$^1$ \hspace{0.2cm}
  Jinwoo Seo$^2$ \hspace{0.2cm}
  Jaewon Cho$^2$ \hspace{0.2cm}
  Jeehye Na$^2$ \\
  \textbf{Yunyang Xiong}$^3$ \hspace{0.2cm}
  \textbf{Youngdae Kim}$^4$ \hspace{0.2cm}
  \textbf{Hyunwoo J. Kim}$^{2}$\thanks{Corresponding author} \\
  $^1$Korea University, $^2$KAIST, $^3$Meta AI, $^4$UNIST\\
  \texttt{allonsy07@korea.ac.kr} \\
  \texttt{\{sjwoo0612, cho35750, jeehyena, hyunwoojkim\}@kaist.ac.kr}\\
  \texttt{yunyang@meta.com} \hspace{0.2cm}
  \texttt{youngdae.kim@unist.ac.kr}
}

\begin{document}

\maketitle

\input{0_Abstract/abstract}
\input{1_Introduction/Introduction}
\input{2_RelatedWorks/RelatedWorks}
\input{3_Preliminary/preliminary}
\input{4_Method/0Intro}
\input{4_Method/1CreditAssignment}
\input{4_Method/2AgentSupervision}
\input{4_Method/3GradientAggregation}
\input{4_Method/4GradientDescent}
\input{5_Experiments/1Experiment_Setting}
\input{5_Experiments/2Analysis}
\input{6_Conclusion/Conclusion}

\begin{ack}
This work was partly supported by the InnoCORE program of the Ministry of Science and ICT (AI Meta-Scientist, N10260110, 30\%); by the Virtual Engineering Platform Project (Grant No. P0022336, 30\%), funded by the Ministry of Trade, Industry \& Energy (MoTIE, South Korea); by the Electronics and Telecommunications Research Institute (ETRI) grant funded by the Korean government (Research on Autonomous Vision Augmentation and Extension Technologies, 26ZR1200, 30\%); and by the Institute of Information \& Communications Technology Planning \& Evaluation (IITP) grant funded by the Korea government (MSIT) (No. RS-2026-25527734, Hyper-scale Industrial AI Research Support (R\&D) Program, Development of industry-specified multimodal hyperscale foundation model based technologies for manufacturing, predictive maintenance, and process optimization, 10\%). The authors declare no competing interests beyond the affiliations and funding sources disclosed in this paper.
\end{ack}

\bibliographystyle{unsrt} 
\bibliography{Reference}

\newpage
\appendix
\input{Appendix/OptimizedPrompts}
\input{Appendix/MetaPrompt}

\input{Appendix/ComputeAnalysis}
\input{Appendix/ResourceMatchedOptimization}

\input{Appendix/SemanticClusteringStability}
\input{Appendix/EmbeddingClusteringComparison}
\input{Appendix/CycleExperiment}
\input{Appendix/AbstractionAblation}
\input{Appendix/GTLeakage}

\input{Appendix/Target_Agent_Attribution}
\input{Appendix/InterventionOrderAnalysis}
\input{Appendix/InterventionOverhead}

\input{Appendix/ValidationCurves}
\input{Appendix/BenchmarkDescriptions}
\input{Appendix/Implementation}
\input{Appendix/BroaderImpactsandLimitations}

\end{document}

%% file: 0_Abstract/abstract.tex
\begin{abstract}
Large language model (LLM)-based multi-agent systems (MAS) achieve strong performance by employing specialized multiple agents, yet their performance depends on the prompt design of each agent.
For MAS prompt optimization, textual gradient methods that guide prompt updates using natural-language feedback have emerged as a leading paradigm.
In this paper, we identify limitations in two stages of existing textual gradient approaches: gradient extraction and gradient aggregation.
In gradient extraction, previous works select a prompt to update without verifying whether modifying it resolves the failure, and derive gradients without agent-level supervision over the corresponding agent’s intermediate output.
In gradient aggregation, individual gradients are randomly grouped and concatenated, often mixing unrelated failure modes and producing prompts that fail to generalize.
To address these limitations, we propose \textbf{AgentGrad}, a prompt optimization framework for multi-agent systems based on sequential intervention and semantic textual gradient abstraction.
For each failure, sequential intervention injects a hint into one agent at a time to identify the target agent whose modification resolves the failure.
The modified output of the target agent then serves as agent-level supervision for extracting a fine-grained gradient.
Semantic textual gradient abstraction clusters semantically similar gradients to prevent mixing unrelated failure modes, and abstracts each cluster into a generalized gradient that captures the shared corrective pattern.
Experimental results show that AgentGrad achieves state-of-the-art performance across five MAS benchmarks while reducing wall-clock optimization time by $2.5\times$ and optimization cost by 21.8\% on average compared to the next-best baselines.
Our code is available at \href{https://github.com/mlvlab/AgentGrad}{https://github.com/mlvlab/AgentGrad}.
\end{abstract}

%% file: 1_Introduction/Introduction.tex
\section{Introduction}
\label{sec:intro}
Recent advances in large language models (LLMs) have enabled the development of multi-agent systems (MAS), where multiple LLM-powered agents interact to solve complex tasks~\citep{wu2024autogen,li2023camel,hong2023metagpt,qian2024chatdev,lin2026squidagent}.
A key advantage of MAS is that they decompose difficult problems into subtasks, allowing different agents to contribute complementary capabilities~\citep{wu2024autogen,li2023camel,hong2023metagpt}.
Such systems have shown strong performance across a wide range of challenging settings, including multi-step reasoning, planning, and information synthesis~\citep{wu2024autogen,hong2023metagpt,qian2024chatdev,islam2024mapcoder,lei2024macm}.
The behavior of each agent is governed by its input prompt, making prompt design critical to system performance~\citep{khattab2023dspy,opsahl2024optimizing}.
Motivated by this, recent works have explored automatic prompt optimization for MAS showing that refining agent prompts can substantially improve system performance~\citep{khattab2023dspy,opsahl2024optimizing,yuksekgonul2024textgrad,agrawal2025gepa,pryzant2023automatic,yang2023large,zhou2022large}. 
Among these, textual gradient methods, which employ natural-language feedback to iteratively refine prompts, have emerged as a leading paradigm~\citep{yuksekgonul2024textgrad,agrawal2025gepa,pryzant2023automatic,sharma2026modular,cui2024introducing}.

In this paper, we identify systematic limitations in two stages of existing textual gradient approaches for MAS: \emph{textual gradient extraction} and \emph{textual gradient aggregation}.
For gradient extraction, previous works exhibit two issues. 
First, the target prompt is selected without verifying whether modifying an individual prompt can resolve the failure.
Existing methods either update all agent prompts simultaneously at substantial cost, or apply round-robin selection without testing which agent can repair the failure.
Second, the gradient is derived without direct agent-level supervision over individual agents' intermediate outputs.
While agent-level supervision provides a direct update signal, it is unavailable in most MAS settings.
In gradient aggregation, individual textual gradients are randomly grouped and directly concatenated. 
It often mixes unrelated failure modes and produces prompt that fails to generalize.

\input{1_Introduction/ConceptFigure}
To address these challenges, we propose \textbf{AgentGrad}, a prompt optimization framework based on \emph{sequential intervention} and \emph{semantic textual gradient abstraction}.
Sequential intervention addresses both limitations in the gradient extraction stage~\citep{zhang2025agent,zhang2025agentracer,wang2026flat}.
Specifically, AgentGrad applies sequential interventions (\textit{e.g.,} hint injection) to individual agents, identifying the target agent whose correction resolves the failure~\citep{zhang2025agent,zhang2025agentracer,wang2026flat,in2026rethinking,chen2026seeing}. 
The intervention-adjusted output then serves as an agent-level pseudo-label to yield a fine-grained textual gradient~\citep{yuksekgonul2024textgrad,pryzant2023automatic,li2024learning,jiao2024preference,lin2024prompt}.
Semantic textual gradient abstraction improves the gradient aggregation stage.
It groups sample-level gradients that share a corrective pattern into semantic minibatches, and abstracts each minibatch into a single generalized textual gradient that captures the shared pattern~\citep{yuksekgonul2024textgrad,agrawal2025gepa,pryzant2023automatic,sharma2026modular,cui2024introducing,ding2025scaling}.

We evaluate AgentGrad on five MAS benchmarks spanning diverse task types: multi-hop QA~\citep{yang2018hotpotqa}, claim verification~\citep{jiang2020hover}, instruction following~\citep{pyatkin2025generalizing}, privacy-conscious delegation~\citep{siyan2024papillon}, and math reasoning~\citep{hendrycks2021measuring}, using both open-source~\citep{yang2025qwen3} and proprietary~\citep{openai2026gpt5mini} backbones.
Across all benchmarks, AgentGrad consistently outperforms strong recent prompt optimization baselines, while substantially reducing wall-clock optimization time and API cost.

Our contributions are summarized as follows:
\begin{itemize}
    \item We propose \textbf{AgentGrad}, a prompt optimization framework for MAS that addresses limitations in two stages of existing textual gradient approaches: gradient extraction and gradient aggregation.
    \item We introduce \emph{sequential intervention}, a mechanism that resolves two limitations in gradient extraction stage: it identifies the target agent whose correction resolves the failure, and produces an agent-level pseudo-label that supplies fine-grained supervision for gradient extraction.
    \item We introduce \emph{semantic textual gradient abstraction}, which groups sample-level gradients into semantic minibatches sharing a corrective pattern and abstracts each cluster into a single textual gradient with improved generalizability.
    \item Across five MAS benchmarks, AgentGrad achieves state-of-the-art performance while reducing wall-clock optimization time by $2.5\times$ and API cost by 21.8\% on average compared to the next-best baselines.
\end{itemize}

%% file: 1_Introduction/ConceptFigure.tex
\begin{figure}[t]
    \centering
    \includegraphics[width=\textwidth]{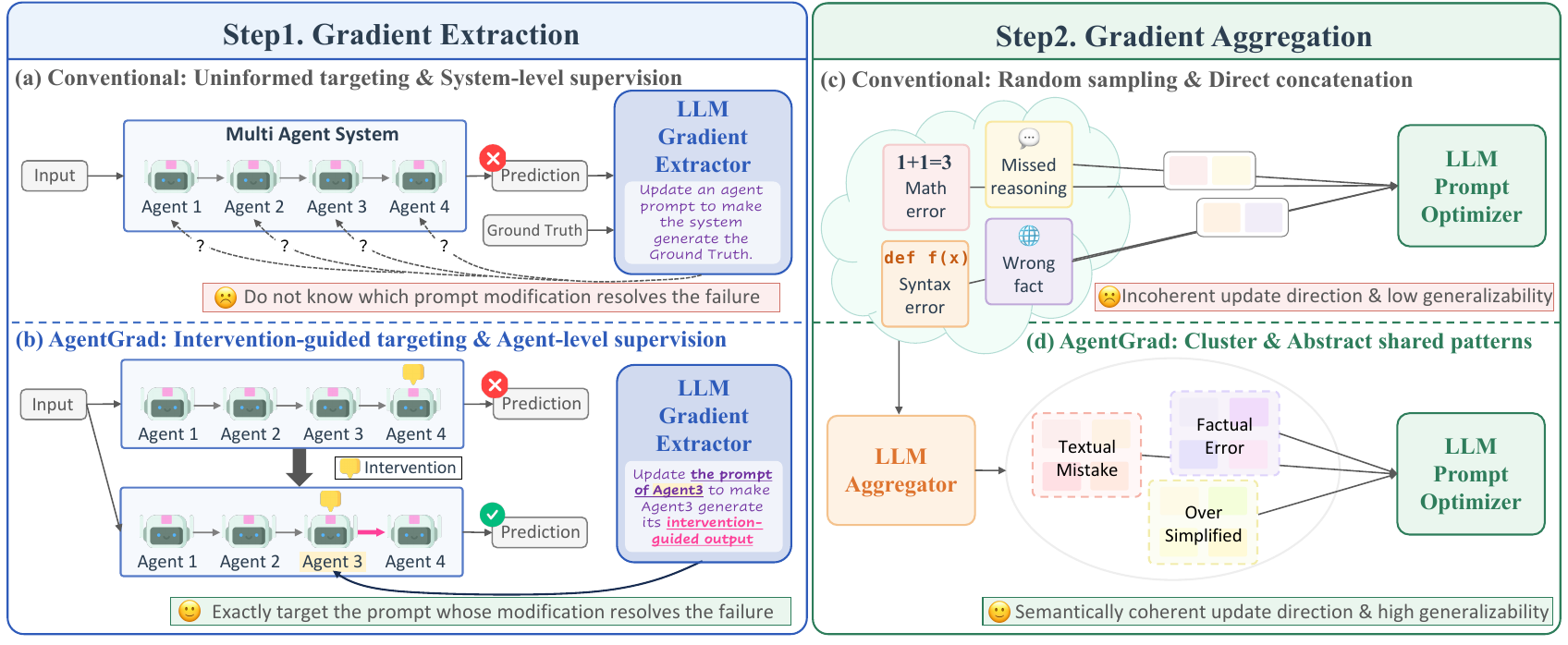}
    \caption{\textbf{Comparison of conventional textual gradient approaches and AgentGrad.} In gradient extraction, conventional approaches (a) select prompts to update without identifying whose modification resolves the failure and extract gradients without agent-level intermediate supervision, while AgentGrad (b) identifies the target agent via sequential intervention and extracts gradients using the intervention-adjusted output of the target agent. In gradient aggregation, conventional approaches (c) randomly group gradients causing spurious signals, while AgentGrad (d) clusters gradients by shared patterns and abstracts them into a generalized gradient.}
    \label{fig:concept_fig}
\end{figure}

%% file: 2_RelatedWorks/RelatedWorks.tex
\section{Related Works}

\paragraph{Automatic Prompt Optimization for LLM-Driven Agents}
Automatic prompt optimization (APO) improves LLM-driven agents by refining instructions, demonstrations, and other textual inputs \cite{khattab2023dspy,opsahl2024optimizing,ramnath2025systematic}.
Early APO mainly targeted single-prompt settings through black-box search, edit-based instruction search, or LLM-generated feedback \cite{yang2023large,zhou2022large,chu2025presto,prasad2023grips,guo2023evoprompt}.
ProTeGi~\cite{pryzant2023automatic} introduced textual gradients by treating natural-language critiques of failed examples as gradient-like directions for prompt revision.
Recent work extends APO from isolated prompts to compound agent systems \cite{khattab2023dspy,opsahl2024optimizing,yuksekgonul2024textgrad,agrawal2025gepa}.
MIPRO~\cite{opsahl2024optimizing} jointly searches instructions and demonstrations for multi-stage LM programs with Bayesian optimization~\cite{frazier2018tutorial, chu2024inversion, lee2023advancing}.
TextGrad~\cite{yuksekgonul2024textgrad} extends textual gradients to compound LLM systems by propagating natural-language feedback across multiple LLM components, analogous to backpropagation.
GEPA~\cite{agrawal2025gepa} combines trajectory-level reflection with evolutionary prompt search.
These methods provide strong APO signals, typically through task-level feedback or trajectory-level reflection \cite{opsahl2024optimizing,yuksekgonul2024textgrad,agrawal2025gepa}.
However, intermediate agent behaviors and interactions could provide finer-grained signals for multi-agent prompt optimization, yet this direction remains underexplored.

\paragraph{Failure Attribution in Multi-Agent Systems}
Failure attribution in multi-agent systems is challenging since system-level errors can emerge from interactions among multiple agents \cite{zhang2025agent,in2026rethinking,chen2026seeing}.
This creates a credit-assignment problem: a failed final output gives limited evidence about which agent or step is associated with the failure \cite{zhang2025agent,nagpal2025leveraging}.
Recent benchmarks and diagnostic methods formalize this problem by localizing failures within multi-agent trajectories \cite{zhang2025agent,zhang2025agentracer,wang2026flat,in2026rethinking,chen2026seeing,yu2025correct}.
Intervention-driven debugging systems further study these failures by editing, replaying, or perturbing agent executions to test how alternative behaviors affect the outcome \cite{zhang2025agentracer,epperson2025interactive,ma2025dover}.
Together, these works establish intervention as a practical tool for failure localization, attribution validation, and multi-agent debugging \cite{epperson2025interactive,ma2025dover}.
AgentGrad extends this line of work from failure analysis to prompt optimization.
It treats the corrected behavior revealed by intervention as agent-level supervision, turning attribution evidence into a localized target for updating the corresponding agent prompt.

\paragraph{Self-Generated Supervision}
Self-generated supervision studies how LLMs can produce rationales, feedback, or reflections as optimization signals \cite{zelikman2022star,shinn2023reflexion,madaan2023self}.
STaR bootstraps rationales to improve reasoning \cite{zelikman2022star}.
Reflexion and Self-Refine use model-generated feedback to refine later attempts, revised outputs, or future task behavior \cite{shinn2023reflexion,madaan2023self}.
These methods establish LLM-generated supervision as an effective optimization source without dense human annotations.
Prior work primarily uses such supervision to improve reasoning trajectories, iterative outputs, or training objectives \cite{jiao2024preference,zelikman2022star,shinn2023reflexion,madaan2023self}, rather than to derive localized prompt updates for individual agents.
AgentGrad redirects this supervision toward prompt-level optimization in multi-agent systems, translating agent-level behavioral correction into supervision for revising agent prompts.

%% file: 3_Preliminary/Preliminary.tex
\section{Preliminaries}
\label{sec:prelim}
\paragraph{Prompt Optimization for MAS}
Let $\Pi$ denote an MAS composed of $N$ LLM-based agents $(\pi^1, \ldots, \pi^N )$. 
Given an input $x$, $\Pi$ generates output $\hat{y} = \Pi(x; \mathcal{P})$, where $\mathcal{P} = ( p^1, \ldots, p^N)$ denotes the collection of agent prompts. 
Given a reward function $r: \hat{\mathcal{Y}} \times \mathcal{Y} \to [0,1]$, prompt optimization finds $\mathcal{P}^*$ that maximizes the expected reward by exploring candidate prompts and validating improvement using a training set $\mathcal{D}_\text{train}$ and a validation set $\mathcal{D}_\text{val}$, respectively.
Since rollouts---running $\Pi$ on an input followed by evaluation under reward function $r$---are computationally expensive, prompt optimization is typically formulated as finding the best solution within a budget of $B$ rollouts allowed~\cite{opsahl2024optimizing,agrawal2025gepa}:
\begin{equation}
    \mathcal{P}^* = \arg\max_{\mathcal{P}}\, 
    \mathbb{E}_{(x, y) \sim \mathcal{D}_\text{val}}\, 
    r\!\left(\Pi(x; \mathcal{P}),\, y\right), 
    \quad \text{s.t. } \#\text{rollouts} \leq B.
\end{equation}
The optimized $\mathcal{P}^*$ is then evaluated on a held-out test set $\mathcal{D}_\text{test}$.
For the $i$-th input $x_i$, the system produces a sequence of intermediate inputs and outputs, where $\hat{y}_i^n = \pi^n(x_i^n; p^n)$, $x_i^1 = x_i$, $\hat{y}_i^N = \hat{y}_i$, and $x^n_i$ for $n\ge 2$ is constructed from preceding agents' outputs.

\paragraph{Textual Gradient}
The \emph{textual gradient}~\cite{pryzant2023automatic} is a natural-language analog of the numerical gradient used in gradient-based optimization. 
Following the formulation in TextGrad~\cite{yuksekgonul2024textgrad}, the textual gradient with respect to a prompt $p$ is defined as 
\begin{equation}
\frac{\partial \mathcal{L}}{\partial p} = \text{LLM}_\nabla\!\left(p, \hat{y}, \mathcal{L}\right),
\end{equation}
where $\mathcal{L}$ is an objective, which may be either a non-differentiable function or a natural-language description of the failure, $\hat{y}$ is the prompt-conditioned output, and $\text{LLM}_\nabla$ is an LLM-based gradient extractor that produces a natural-language critique describing how $p$ should be modified to improve $\mathcal{L}$. 
A separate prompt optimizer LLM then aggregates sample-level textual gradients to produce an updated prompt. 
This extraction–aggregation procedure underlies a family of textual gradient methods~\cite{yuksekgonul2024textgrad,sharma2026modular,cui2024introducing}.
Some recent works refer to this signal under different names. 
GEPA, for instance, calls it natural-language feedback~\cite{agrawal2025gepa}, but these methods share the same abstraction: a natural language description of how a prompt should be updated.
We adopt \emph{textual gradient} throughout this work to denote this signal.

%% file: 4_Method/0Intro.tex
\section{Method}
\label{sec:method}
In this section, we present \textbf{AgentGrad}, a prompt optimization framework for multi-agent systems based on \emph{sequential intervention} and \emph{semantic textual gradient abstraction}.
For the gradient extraction stage, we introduce sequential intervention, a mechanism that resolves the two limitations of this stage. 
Sequential intervention identifies the agent whose correction resolves a system-level failure (Section~\ref{main_sec:CreditAssignment}). 
The intervention-induced output then serves as an agent-level pseudo-label that supplies fine-grained supervision for textual gradient extraction (Section~\ref{main_sec:AgentSupervision}).
For the gradient aggregation stage, we introduce semantic textual gradient abstraction (Section~\ref{main_sec:GradientAggregation}), which groups sample-level gradients into semantic minibatches sharing a corrective pattern and abstracts each cluster into a single textual gradient with improved generalizability.
We provide the pseudocode of AgentGrad in algorithm~\ref{alg:agentgrad}.
\input{4_Method/MainFigure}

%% file: 4_Method/MainFigure.tex
\begin{figure*}[!t]
\centering
\begin{minipage}[t]{0.48\textwidth}
\vspace{0pt}
\centering
\includegraphics[width=\linewidth,height=0.47\textheight,keepaspectratio]{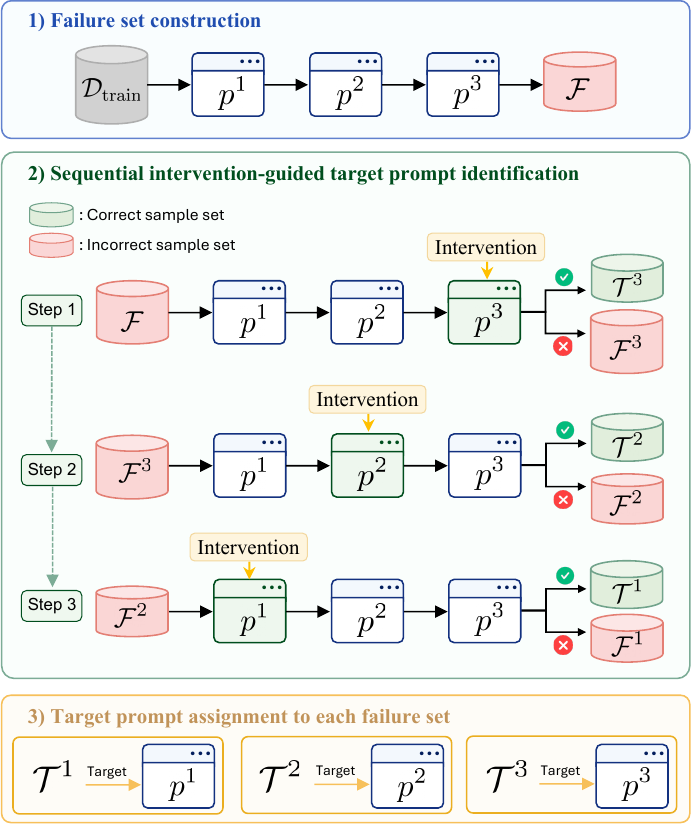}
\captionof{figure}{\textbf{Intervention-Guided Target Identification.} 
AgentGrad first executes the current prompt set on $\mathcal{D}_{\text{train}}$ to obtain the failed examples $\mathcal{F}$. 
It then applies interventions to each agent in reverse execution order, progressively separating the unresolved failures into corrected subsets $\mathcal{T}^{n}$ and remaining failure sets $\mathcal{F}^{n}$. 
Here, $\mathcal{T}^{n}$ denotes the subset of failures for which the $n$-th agent is identified as the target, namely those resolved by intervening on prompt $p^{n}$.}
\label{fig:main_fig}
\end{minipage}
\hfill
\begin{minipage}[t]{0.48\textwidth}
\vspace{-1.1em}
\begin{algorithm}[H]
\footnotesize
\caption{AgentGrad}
\label{alg:agentgrad}
\begin{algorithmic}[1]
\Require MAS $\Pi$, prompts $\mathcal{P}$, train/val sets $\mathcal{D}_{\text{train}},\mathcal{D}_{\text{val}}$, reward $r$, max reward $r_{\max}$, budget $B$, Hint $\mathcal H$
\State $r_{\mathcal{P}}(x,y) := r(\Pi(x;\mathcal{P}),y)$
\State $R_{\mathcal{D}}(\mathcal{P}) := \mathbb{E}_{(x,y)\sim\mathcal{D}}[r_{\mathcal{P}}(x,y)]$
\State $\text{Fail}(\mathcal{P};\mathcal{S}) :=
\{(x,y)\in\mathcal{S}: r_{\mathcal{P}}(x,y)<r_{\max}\}$
\State $\text{Fail}^{(n,\mathcal{H})}$ and $r_{\mathcal{P}}^{(n,\mathcal{H})}$ denote $\text{Fail}$ and $r_{\mathcal{P}}$ after intervening on $\pi^n$ with hint $\mathcal{H}$

\While{rollout budget is not exhausted}
    \State $\mathcal{F} \leftarrow \text{Fail}(\mathcal{P};\mathcal{D}_{\text{train}})$
    \State $\Omega^n \leftarrow \emptyset , \mathcal{F}^{N+1} \leftarrow \mathcal{F}\quad \forall n\in\{1,\ldots,N\}$

    \For{$n=N,\ldots,1$}
        \State $\mathcal{F}^n \leftarrow \text{Fail}^{(n,\mathcal{H})}(\mathcal{P};\mathcal{F}^{n+1})$
        \State $\mathcal{T}^n \leftarrow \mathcal{F}^{n+1} \setminus \mathcal{F}^n$

        \For{each $(x_i,y_i)\in\mathcal{T}^n$}
            \State Obtain $(x_i^n,\hat{y}_i^n,\tilde{y}_i^n)$
            \State $\delta_i^n \leftarrow
            \text{LLM}_{\nabla}(p^n,x_i^n,\hat{y}_i^n,\tilde{y}_i^n)$
            \State $\Omega^n \leftarrow \Omega^n\cup\{\delta_i^n\}$
        \EndFor
    \EndFor

    \For{each $n$}
        \State $\{\bar{\delta}_j^n\}_{j=1}^{M_n} \leftarrow \text{LLM}_{\text{Aggregator}}(\Omega^n)$
            \Comment{with semantic minibatch $\mathcal{D}_j^n$ for each $\bar{\delta}_j^n$}
    \EndFor

    \For{each $n,j$ in decreasing order of $|\mathcal{D}_j^n|$}
        \State $p_{\text{new}}^n \leftarrow
        \text{LLM}_{\text{PromptOptimizer}}(p^n,\bar{\delta}_j^n)$
        \State $\mathcal{P}_{\text{new}} \leftarrow \mathcal{P}$ with $p^n$ replaced by $p_{\text{new}}^n$
        \State \textbf{if} $R_{\mathcal{D}_j^n}(\mathcal{P}_{\text{new}}) > R_{\mathcal{D}_j^n}(\mathcal{P})$ \textbf{then}
        \State \quad \ \ \textbf{if} $R_{\mathcal{D}_{\text{val}}}(\mathcal{P}_{\text{new}}) > R_{\mathcal{D}_{\text{val}}}(\mathcal{P})$ \textbf{then}
        \State \quad \ \ \quad \ \ $\mathcal{P} \leftarrow \mathcal{P}_{\text{new}}$
        \State \quad  \ \ \textbf{end if}
        \State \textbf{end if}
    \EndFor
\EndWhile

\State \Return $\mathcal{P}$
\end{algorithmic}
\end{algorithm}
\end{minipage}

\end{figure*}

%% file: 4_Method/1CreditAssignment.tex
\subsection{Target Prompt Identification via Sequential Intervention}
\label{main_sec:CreditAssignment}
Our goal is to select the target prompt to update.
We recast credit assignment for target prompt selection as identifying an agent whose corrected behavior is sufficient to resolve a failure, rather than attributing responsibility for that failure.
To this end, we identify the target prompt via sequential intervention, which modifies one agent at a time to verify whether its correction resolves the failure.
An intervention modifies an agent's behavior by injecting a hint into its prompt, guiding it toward a corrected intermediate output that leads the system to generate the correct output.
Figure~\ref{fig:main_fig} illustrates the overall procedure of the sequential intervention.

\paragraph{Sequential Intervention.}
Given the current prompt set $\mathcal{P}$ of $N$ agents, let $\mathcal{F} = \{(x_i, y_i) \mid r(\Pi(x_i; \mathcal{P}), y_i) < r_{\max}\}$ denote the set of failures from the training set $\mathcal{D}_\text{train}$, where $r$ is a reward function and $r_{\max}$ is the maximum reward value.
For each failure $(x_i, y_i) \in \mathcal{F}$, we apply interventions one agent at a time in reverse execution order to identify the agent whose correction resolves it. 
We adopt reverse execution order by analogy with the backward pass in backpropagation, starting from the final agent and proceeding upstream.
Intervening on a downstream agent also allows us to reuse the cached outputs of preceding agents and re-execute only the intervened agent and its downstream agents.
Starting with $\mathcal{F}^{N+1} = \mathcal{F}$, we iterate over $n=N,\ldots,1$, injecting hint $\mathcal{H}$ into agent $\pi^n$ by appending it to the agent's current prompt.
The appended hint guides the agent toward a corrected output; we denote the intervened MAS as $\Pi^{(n, \mathcal{H})}$.
We define $\mathcal{T}^n$ as the subset of $\mathcal{F}^{n+1}$ resolved by intervening on $\pi^n$; for this subset, $\pi^n$ is identified as the target agent:
\begin{equation}
    \mathcal{T}^n = \big\{ (x_i, y_i) \in \mathcal{F}^{n+1} \;\big|\; 
    r\!\left(\Pi^{(n, \mathcal{H})}(x_i; \mathcal{P}), y_i\right) = r_{\max} 
    \big\},
\end{equation}
where $\mathcal{F}^{n+1}$ denotes the failures still unresolved before intervening on agent $\pi^n$.
The resolved failures $\mathcal{T}^n$ are then removed from the unresolved set ($\mathcal{F}^{n} = \mathcal{F}^{n+1} \setminus \mathcal{T}^n$), and the procedure continues with agent $\pi^{n-1}$ if $n>1$.
Failures that remain unresolved after interventions have been applied to all agents individually are treated as hard cases and excluded from the current training round.
They are not permanently discarded: since $\mathcal{F}$ is reconstructed from $\mathcal{D}_\text{train}$ at the start of every round, these cases are revisited under the updated prompt set.

\paragraph{Hint Construction.}
The hint $\mathcal{H}$ is designed to provide the necessary guidance to direct an agent toward correct behavior.
We construct $\mathcal{H}$ from the ground-truth $y_i$ or from constraints that the final output must satisfy, together with auxiliary context such as descriptions of the dataset, the MAS, and each agent role.
These components are commonly used in prior prompt optimization methods~\cite{khattab2023dspy}.
All compared methods have access to the same training-set ground-truth information.
We emphasize that $\mathcal{H}$ is used \emph{only at training time}; the optimized prompts are deployed without any hint injection at inference time.

%% file: 4_Method/2AgentSupervision.tex
\subsection{Textual Gradient Extraction via Agent-Level Supervision}
\label{main_sec:AgentSupervision}
In this section, we describe textual gradient extraction using the target agent's intervention-induced output as an agent-level pseudo-label.
For each target agent $\pi^n$ and each failure sample $(x_i, y_i)\in \mathcal{T}^n$ attributed to it (Section~\ref{main_sec:CreditAssignment}), we have two outputs under the same input $x_i^n$: the original output $\hat{y}_i^n$ produced during the failed execution, and the corrected output $\tilde{y}_i^n$ obtained by injecting hint $\mathcal{H}$ during sequential intervention:
\begin{equation}
    \hat{y}_i^n = \pi^n(x_i^n; p^n), \quad \tilde{y}_i^n = \pi^n(x_i^n; p^n, \mathcal{H}).
\end{equation}
Since both $\hat{y}_i^n$ and $\tilde{y}_i^n$ are generated under the same input context $x_i^n$, their difference isolates the behavioral change induced by the intervention.
We therefore interpret $\tilde{y}_i^n$ as an agent-level pseudo-label specifying how $\pi^n$ should behave under $x_i^n$.
To extract a textual gradient, AgentGrad uses this pseudo-label as input to a gradient extractor LLM:
\begin{equation}
    \delta_i^n = \mathrm{LLM}_{\nabla}\!\left(
    p^n, x_i^n, \hat{y}_i^n, \tilde{y}_i^n\right),
\end{equation}
which captures how $p^n$ should be modified to produce $\tilde{y}_i^n$ instead of $\hat{y}_i^n$.
We call $\delta_i^n$ a sample-level textual gradient, since it is derived from a single failure sample $(x_i, y_i)$ and describes the correction required for that instance alone.
Unlike standard textual gradient methods which require an explicit $\mathcal{L}$ derived from comparing the system-level output against the ground-truth, our gradient requires no explicit loss. 
The contrast between $\hat{y}_i^n$ and $\tilde{y}_i^n$ implicitly provides agent-level supervision.
This agent-level supervision yields fine-grained update signals for the target agent.

%% file: 4_Method/3GradientAggregation.tex
\subsection{Semantic Textual Gradient Abstraction}
\label{main_sec:GradientAggregation}
In this section, we introduce semantic textual gradient abstraction, which clusters semantically similar sample-level gradients into semantic minibatches and abstracts each minibatch into a single generalized textual gradient.
Standard textual gradient methods aggregate sample-level gradients from random minibatches. 
Such minibatches often mix gradients from unrelated failure modes leaving the prompt optimizer without a coherent update direction.
In contrast, semantic minibatches contain failures that share a common corrective pattern, providing a coherent update direction.

For each agent $\pi^n$, let $\Omega^n = \{\delta_i^n\}_{(x_i, y_i)\in \mathcal{T}^n}$ denote the set of sample-level textual gradients extracted in Section~\ref{main_sec:AgentSupervision} over all failures attributed to $\pi^{n}$.
We employ an aggregator LLM that clusters $\Omega^n$ into semantic minibatches and abstracts each minibatch into a generalized textual gradient:
\begin{equation}
    \{\bar{\delta}^n_j\}_{j=1}^{M_n} = 
    \mathrm{LLM}_{\text{Aggregator}}(\Omega^n),
\end{equation}
where $\bar{\delta}^n_j$ denotes the $j$-th generalized gradient for agent $\pi^n$ and $M_n$ is the number of resulting clusters, which is determined by the aggregator LLM. 
We use a separate index $j$ to distinguish generalized gradients from sample-level gradients indexed by $i$.
The aggregator LLM performs two coupled steps within a single call: clustering $\Omega^n$ into semantically coherent groups, and abstracting each group into a generalized gradient that captures its shared corrective pattern.

\paragraph{Clustering into Semantic Minibatches.}
The aggregator LLM groups semantically similar sample-level gradients in $\Omega^n$. 
Each group induces a semantic minibatch $\mathcal{D}^n_j$: the training failures from which the sample-level gradients in that group were extracted.
The size of each cluster determines the abstraction level: larger clusters yield more general patterns shared across many failures, while smaller clusters yield finer-grained corrections. 
To guide this abstraction level, we provide the aggregator with a soft lower bound on cluster size, which follows a cyclic schedule across optimization iterations (e.g., $5 \to 3 \to 1 \to 5 \to \dots$). 
This schedule encourages the aggregator to alternate between coarse, broadly-shared patterns and finer, more specific corrections over the course of training. 
The lower bound on cluster size is recommended rather than strictly enforced, allowing the aggregator to form smaller clusters when the gradients are too dissimilar to group.

\paragraph{Abstracting into Generalized Gradients.}
For each cluster, the aggregator LLM produces a generalized textual gradient $\bar{\delta}^n_j$ that captures the shared corrective pattern of its semantic minibatch. 
This provides a coherent update direction while reducing reliance on instance-specific details, encouraging prompt updates that generalize beyond individual failure cases.
The resulting generalized gradients $\{\bar{\delta}^n_j\}$ and their semantic minibatches $\{\mathcal{D}^n_j\}$ serve as the primary signal for prompt updates.

%% file: 4_Method/4GradientDescent.tex
\subsection{Prompt Update and Validation}
\label{main_sec:PromptUpdate}
Given the generalized gradients $\{\bar{\delta}_j^n\}$ from Section~\ref{main_sec:GradientAggregation}, we update agent prompts in decreasing order of semantic minibatch size, applying broader, high-influence updates before finer ones.
A prompt optimizer LLM generates a candidate prompt $p_{\text{new}}^n$ using the generalized gradient $\bar{\delta}_j^n$ and the current prompt $p^n$:
\begin{equation}
    p_{\text{new}}^n = \mathrm{LLM}_{\text{PromptOptimizer}}\!\left(
    p^n, \bar{\delta}_j^n\right).
\end{equation}
Each candidate prompt is first evaluated on the semantic minibatch $\mathcal{D}_j^n$; if performance improves, it is then evaluated on the held-out validation set $\mathcal{D}_{\text{val}}$.
If both stages pass, we accept the update by replacing $p^n$ in $\mathcal{P}$ with $p_{\text{new}}^n$; otherwise, we discard it and proceed to the next gradient.

%% file: 5_Experiments/1Experiment_Setting.tex
\section{Experiment}
\label{sec:experiment}

\subsection{Experimental Setup}
\label{sec:exp_setup}
We evaluate AgentGrad against three state-of-the-art prompt optimization algorithms---MIPROv2~\cite{opsahl2024optimizing}, TextGrad~\cite{yuksekgonul2024textgrad}, and GEPA~\cite{agrawal2025gepa}---and a no-optimization baseline across five MAS benchmarks: HotpotQA, HoVer, IFBench, PUPA, and MATH~\cite{yang2018hotpotqa,jiang2020hover,pyatkin2025generalizing,siyan2024papillon,hendrycks2021measuring}.
For HotpotQA, HoVer, PUPA, and IFBench, we adopt the multi-agent systems, data splits, and reward functions from~\cite{agrawal2025gepa}; for MATH, we adopt from~\cite{lei2024macm}.
We evaluate two LLM backbones, GPT-5-mini and Qwen3-8B~\cite{yang2025qwen3}, where the same backbone serves as both the task LLM and all optimizer components across all optimization algorithms.

\subsection{Main Results}
\label{sec:main_results}
\input{tables/main_table_gpt}
\input{tables/main_table_qwen}

AgentGrad achieves state-of-the-art average performance across five MAS benchmarks on both backbone settings.
With GPT-5-mini in Table~\ref{tab:main_gpt}, AgentGrad outperforms all baselines on every benchmark, achieving an average improvement of $+11.76$ points over the no-optimization baseline, with the largest margins on HotpotQA ($73.89$ vs.\ $68.33$ for GEPA) and PUPA ($95.17$ vs.\ $91.87$).
In Table~\ref{tab:main_qwen}, AgentGrad with Qwen3-8B achieves the largest average improvement of $+9.67$ points over the no-optimization baseline, surpassing all baselines.
The consistent gains across both proprietary and open-source models, spanning multi-hop QA, claim verification, instruction following, privacy-conscious delegation, and math reasoning, demonstrate that sequential intervention and semantic textual gradient abstraction generalize across diverse task types and agent configurations.

%% file: tables/main_table_gpt.tex
\begin{table}[!t]
\caption{\textbf{Main results on five MAS benchmarks with GPT-5-mini.} 
We report the mean $\pm$ standard error over three random seeds. \textbf{Bold} indicates the best result.}
\label{tab:main_gpt}
\definecolor{oursbg}{RGB}{234,242,248}
\centering
\small
\setlength{\tabcolsep}{4pt}
\begin{tabular}{lcccccc}
\toprule
\textbf{GPT-5-mini} & \textbf{HotpotQA} & \textbf{HoVer} & \textbf{PUPA} & \textbf{IFBench} & \textbf{MATH} & \textbf{Improvement}\\
\midrule
Baseline (No PO) & 46.33 {\scriptsize $\pm$ 0.69} & 58.11 {\scriptsize $\pm$ 1.01} & 84.74 {\scriptsize $\pm$ 0.33} & 73.07 {\scriptsize $\pm$ 0.60} & 76.48 {\scriptsize $\pm$ 0.91} & - \\
MIPROv2          & 59.00 {\scriptsize $\pm$ 1.66} & 62.89 {\scriptsize $\pm$ 1.34} & 88.33 {\scriptsize $\pm$ 2.13} & 73.70 {\scriptsize $\pm$ 0.86} & 83.13 {\scriptsize $\pm$ 1.72} & +5.66 \\
TextGrad         & 67.89 {\scriptsize $\pm$ 1.31} & 63.22 {\scriptsize $\pm$ 1.47} & 89.72 {\scriptsize $\pm$ 2.43} & 73.07 {\scriptsize $\pm$ 0.60} & 76.48 {\scriptsize $\pm$ 0.91} & +6.33 \\
GEPA             & 68.33 {\scriptsize $\pm$ 1.55} & 63.11 {\scriptsize $\pm$ 1.90} & 91.87 {\scriptsize $\pm$ 1.55} & 75.23 {\scriptsize $\pm$ 0.32} & 86.37 {\scriptsize $\pm$ 0.50} & +9.24 \\
\rowcolor{oursbg}
\textbf{AgentGrad (Ours)} & \textbf{73.89} {\scriptsize $\pm$ 1.09} & \textbf{64.78} {\scriptsize $\pm$ 1.44} & \textbf{95.17} {\scriptsize $\pm$ 0.49} & \textbf{76.08} {\scriptsize $\pm$ 0.45} & \textbf{87.62} {\scriptsize $\pm$ 0.09} & \textbf{+11.76} \\
\bottomrule
\end{tabular}
\end{table}

%% file: tables/main_table_qwen.tex
\begin{table}[!t]
\caption{\textbf{Main results on five MAS benchmarks with Qwen3-8B.} We report the mean $\pm$ standard error over three random seeds. \textbf{Bold} indicates the best result.}
\label{tab:main_qwen}
\definecolor{oursbg}{RGB}{234,242,248}
\centering
\small
\setlength{\tabcolsep}{4pt}
\begin{tabular}{lcccccc}
\toprule
\textbf{Qwen3-8B} & \textbf{HotpotQA} & \textbf{HoVer} & \textbf{PUPA} & \textbf{IFBench} & \textbf{MATH} & \textbf{Improvement}\\
\midrule
Baseline (No PO) & 41.33 {\scriptsize $\pm$ 0.84} & 36.67 {\scriptsize $\pm$ 1.02} & 80.87 {\scriptsize $\pm$ 0.06} & 40.82 {\scriptsize $\pm$ 1.99} & 83.24 {\scriptsize $\pm$ 0.38} & - \\
MIPROv2          & 58.33 {\scriptsize $\pm$ 2.37} & 45.44 {\scriptsize $\pm$ 0.68} & 85.76 {\scriptsize $\pm$ 2.86} & 40.08 {\scriptsize $\pm$ 2.21} & 84.68 {\scriptsize $\pm$ 0.92}& +6.27 \\
TextGrad         & 50.86 {\scriptsize $\pm$ 5.36} & 51.44 {\scriptsize $\pm$ 0.78} & 84.50 {\scriptsize $\pm$ 2.31} & \textbf{42.52} {\scriptsize $\pm$ 0.45} & 83.90 {\scriptsize $\pm$ 0.67} & +6.06 \\
GEPA             & 57.33 {\scriptsize $\pm$ 2.54} & 50.11 {\scriptsize $\pm$ 1.60} & 91.03 {\scriptsize $\pm$ 1.71} & 37.53 {\scriptsize $\pm$ 1.62} & 85.05 {\scriptsize $\pm$ 0.48} & +7.62\\
\rowcolor{oursbg}
\textbf{AgentGrad (Ours)} & \textbf{60.45} {\scriptsize $\pm$ 1.68} & \textbf{52.11} {\scriptsize $\pm$ 1.66} & \textbf{91.51} {\scriptsize $\pm$ 0.72} & 41.42 {\scriptsize $\pm$ 0.99} & \textbf{85.81} {\scriptsize $\pm$ 0.25} & \textbf{+9.67}\\
\bottomrule
\end{tabular}
\end{table}

%% file: 5_Experiments/2Analysis.tex
\subsection{Analysis}
\label{sec:analysis}

\paragraph{Ablation Studies.}
\input{tables/ablation_table}
Table~\ref{tab:ablation} presents an ablation study on HotpotQA and PUPA with GPT-5-mini, progressively adding each component of AgentGrad to a vanilla baseline. 
All three components contribute positively. Target identification (TI) using sequential-intervention alone yields $+1.44$ and $+3.84$ points on HotpotQA and PUPA, showing that identifying the repair target agent already produces more effective gradients. 
Adding agent-level supervision (AS) on top of TI yields a further $+1.56$ and $+2.65$ points, while semantic textual gradient abstraction (STGA) on top of TI contributes $+2.56$ and $+3.55$ points. 
The full model (AgentGrad) achieves the best performance on both benchmarks, confirming that TI, AS, and STGA address complementary aspects of the prompt optimization process.

\paragraph{Optimization Trajectory.}
Figure~\ref{fig:trajectory} shows validation performance as a function of the number of rollouts on HotpotQA (GPT-5-mini). AgentGrad consistently dominates all baselines at every rollout count.
On HotpotQA, AgentGrad achieves approximately $70\%$ by $1{,}000$ rollouts, while GEPA requires over $6{,}000$ rollouts to approach a comparable level, and MIPROv2 and TextGrad plateau below this threshold.
AgentGrad's confidence bands are also notably narrow, reflecting more stable optimization across seeds, and the performance gap is maintained rather than narrowing with additional rollouts.

\paragraph{Wall-clock Time Comparison.}
\input{tables/wallclocktime_table}
\input{tables/cost_table}
Table~\ref{tab:wallclock} reports wall-clock optimization time on five benchmarks with GPT-5-mini.
AgentGrad is the fastest method across all five benchmarks without exception, completing optimization in $136$ minutes on average, $2.5\times$ faster than the next-fastest baseline.
The speedup is most pronounced on HotpotQA ($3.2\times$ over GEPA) and IFBench ($3.0\times$), where AgentGrad finishes in under two hours while GEPA requires nearly six.
Even on MATH, where TextGrad is unusually fast at $126$ minutes, AgentGrad completes in just $88$ minutes.
Notably, AgentGrad achieves these speedups while simultaneously attaining the best task performance (Table~\ref{tab:main_gpt}), demonstrating that the two objectives — optimization quality and efficiency — are not in tension but are jointly improved by agent-level gradient signals.

\paragraph{Optimization Cost Analysis.}
AgentGrad is not only faster but also substantially more cost-efficient than existing prompt optimization methods. 
As shown in Table~\ref{tab:cost}, AgentGrad achieves the lowest average optimization cost of \$27.14, reducing the cost by 21.8\% compared with the next-best baseline. 
Notably, AgentGrad is the most cost-efficient method on four of the five benchmarks, with cost reductions of up to 37.1\% on MATH. 
This advantage is particularly pronounced over TextGrad and MIPROv2, whose average costs are 3.6$\times$ and 2.0$\times$ higher than that of AgentGrad, respectively. 
Together with the substantial wall-clock speedups in Table~\ref{tab:wallclock}, these results demonstrate that AgentGrad improves optimization efficiency along both key dimensions---\emph{time and monetary cost}---while simultaneously achieving the strongest overall task performance (Table~\ref{tab:main_gpt}).

\paragraph{Why AgentGrad Optimizes Faster and Generalizes Better.}
\input{5_Experiments/ratio}
\input{tables/generality}
\input{tables/intervention_coverage}
Figure~\ref{fig:ratios}(a–b) compares minibatch and validation improvement ratios against GEPA and TextGrad, averaged over HotpotQA and PUPA.
AgentGrad achieves a minibatch improvement ratio of $0.72$ versus $0.44$ for TextGrad and $0.28$ for GEPA.
Since validation is triggered only when a candidate improves the minibatch, this higher ratio leads to more frequent evaluations on the larger validation set. 
Together with effective cache reuse, this allows more rollouts to be processed per unit time, helping explain AgentGrad's wall-clock speedup under a fixed rollout budget.
This efficiency does not come at the cost of quality: AgentGrad also achieves the highest validation improvement ratio ($0.27$ vs.\ $0.21$ and $0.14$), indicating that its accepted updates generalize more reliably.
Figure~\ref{fig:ratios}(c–d) isolates each component's contribution.
TI and AS primarily raise the minibatch ratio (from $0.51$ to $0.87$), improving per-sample gradient signal quality, while STGA trades a modest minibatch ratio reduction for a gain in validation ratio, improving generalizability --- yielding a clear division of roles among the three components.

\paragraph{Transferability of Optimized Prompts.}
Table~\ref{tab:benchmark_results} evaluates whether optimized prompts remain effective on an unseen benchmark from the same domain~\cite{ho2020constructing, ma2024ex, siyan2024papillon, zhou2023instruction, he2024olympiadbench}, without any further optimization. 
AgentGrad achieves the best transfer performance on all five target benchmarks compared to strong prompt optimization algorithms.
The margin over the next-best method is largest on 2WikiMultiHopQA (51.22 vs. 44.89 for GEPA) and PUPA-TNB (94.38 vs. 91.51).
Together with the in-domain results (Tables~\ref{tab:main_gpt}, ~\ref{tab:main_qwen}) and the optimization time comparison (Table~\ref{tab:wallclock}), this shows that AgentGrad's gains are not confined to the benchmark it was optimized on: the same prompts remain the strongest on unseen benchmarks within the domain.

\paragraph{Coverage of sequential intervention.}
We assess the coverage of sequential intervention across five benchmarks spanning diverse MAS architectures.
As shown in Table~\ref{tab:intervention_coverage}, sequential intervention resolves 186 of 244 initially failed examples (76.2\%).
Simultaneous intervention resolves four additional cases, increasing coverage to 77.9\%.
This small gap suggests that sequential intervention captures most of the failure-resolution benefit of simultaneous intervention across the evaluated architectures.
Directly using all simultaneously corrected outputs as supervision would entail updating all agent prompts.
Sequential intervention instead identifies an individual repair target and provides its corrected output for each resolved failure, enabling localized prompt updates.

\begin{figure}[t]
    \centering
    \includegraphics[width=0.99\linewidth]{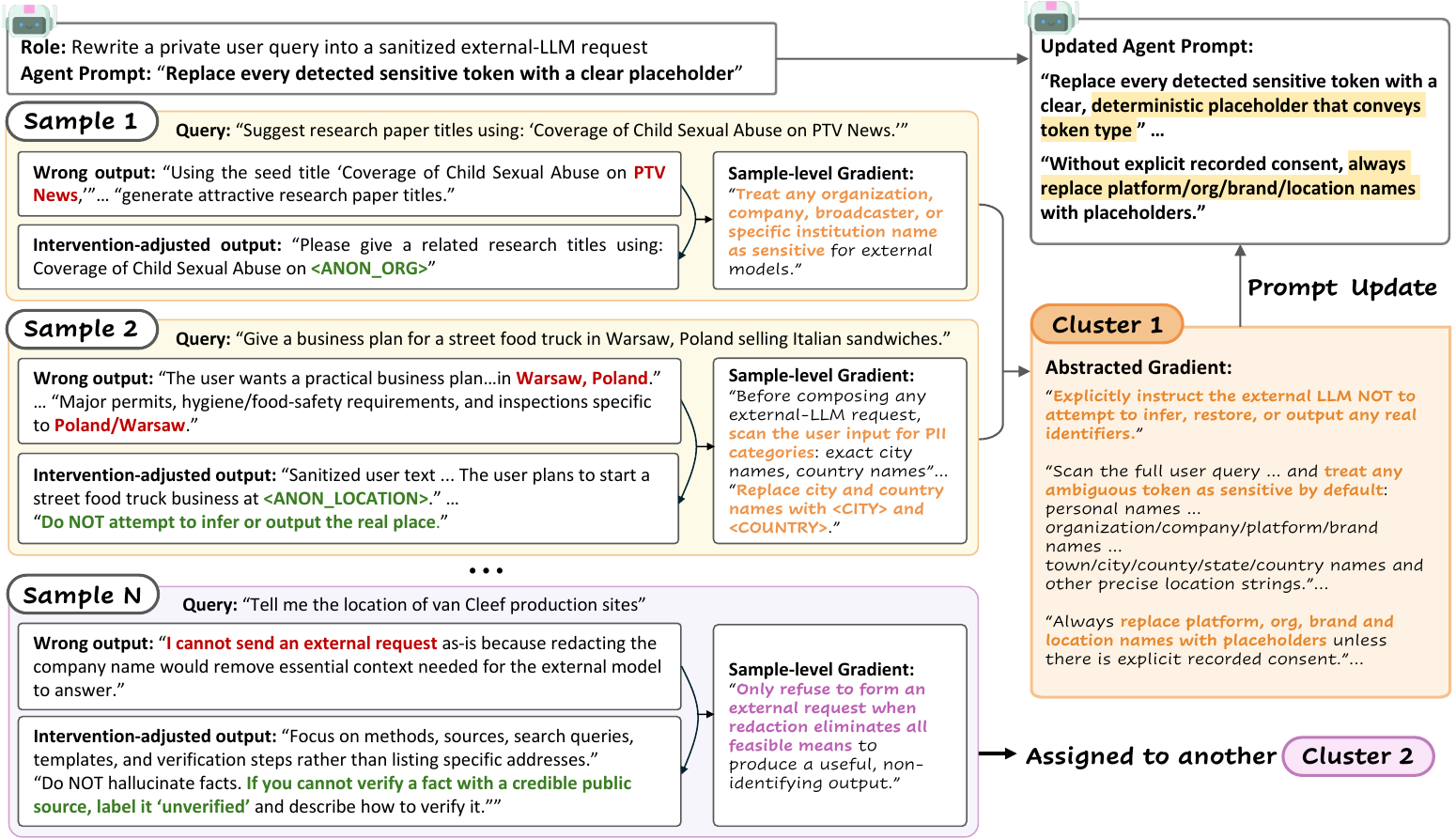}
    \caption{
    \textbf{Qualitative example of gradient extraction, semantic clustering, and abstraction.}
    Failed and intervention-adjusted outputs guide sample-level gradient extraction.
    Gradients addressing identifier leakage (Samples 1 and 2) are clustered and abstracted for prompt updating, while the gradient addressing excessive refusal (Sample N) forms a separate cluster.
    }
    \label{fig:qual_gradient_abstraction}
    \vspace{-15pt}
\end{figure}

\paragraph{Qualitative Results.}
Figure~\ref{fig:qual_gradient_abstraction} illustrates how AgentGrad extracts sample-level gradients using intervention-adjusted outputs (Section~\ref{main_sec:AgentSupervision}), clusters semantically similar gradients, and abstracts them for prompt updating (Section~\ref{main_sec:GradientAggregation}).
In Samples 1 and 2, failed outputs expose an organization name and a location (red), whereas intervention-adjusted outputs replace them with placeholders (green).
Comparing each failed output with its intervention-adjusted counterpart yields a sample-level gradient specifying the required prompt correction.
The gradients from Samples 1 and 2 share a corrective signal concerning identifier leakage and are clustered together, then abstracted into a generalized redaction policy used to update the target agent's prompt.
In contrast, the gradient from Sample N addresses excessive refusal and is assigned to a separate cluster.

%% file: tables/ablation_table.tex
\begin{figure}[t]
\centering

\begin{minipage}[t]{0.50\textwidth}
\centering
\captionof{table}{\textbf{Ablation study of AgentGrad.}
We progressively add intervention-guided target identification (TI), agent-level supervision (AS), and semantic textual gradient abstraction (STGA) to a vanilla baseline.}
\label{tab:ablation}
\definecolor{oursbg}{RGB}{234,242,248}
\small
\resizebox{\linewidth}{!}{%
\begin{tabular}{ccc cc}
\toprule
TI & AS & STGA & HotpotQA & PUPA \\
\midrule
        &        &        & 67.89 $\pm$ 0.80 & 85.74 $\pm$ 1.65 \\
\cmark  &        &        & 69.33 $\pm$ 0.38 & 89.58 $\pm$ 1.89 \\
\cmark  & \cmark &        & 70.89 $\pm$ 0.62 & 92.23 $\pm$ 0.68 \\
\cmark  &        & \cmark & 71.89 $\pm$ 0.29 & 93.13 $\pm$ 0.53 \\
\rowcolor{oursbg}
\cmark  & \cmark & \cmark & \textbf{73.89 $\pm$ 1.09} & \textbf{95.17 $\pm$ 0.49} \\
\bottomrule
\end{tabular}%
}
\end{minipage}
\hfill
\begin{minipage}[t]{0.46\textwidth}
\centering
\vspace{0pt}

\centering
\includegraphics[
    width=\linewidth,
    keepaspectratio
]{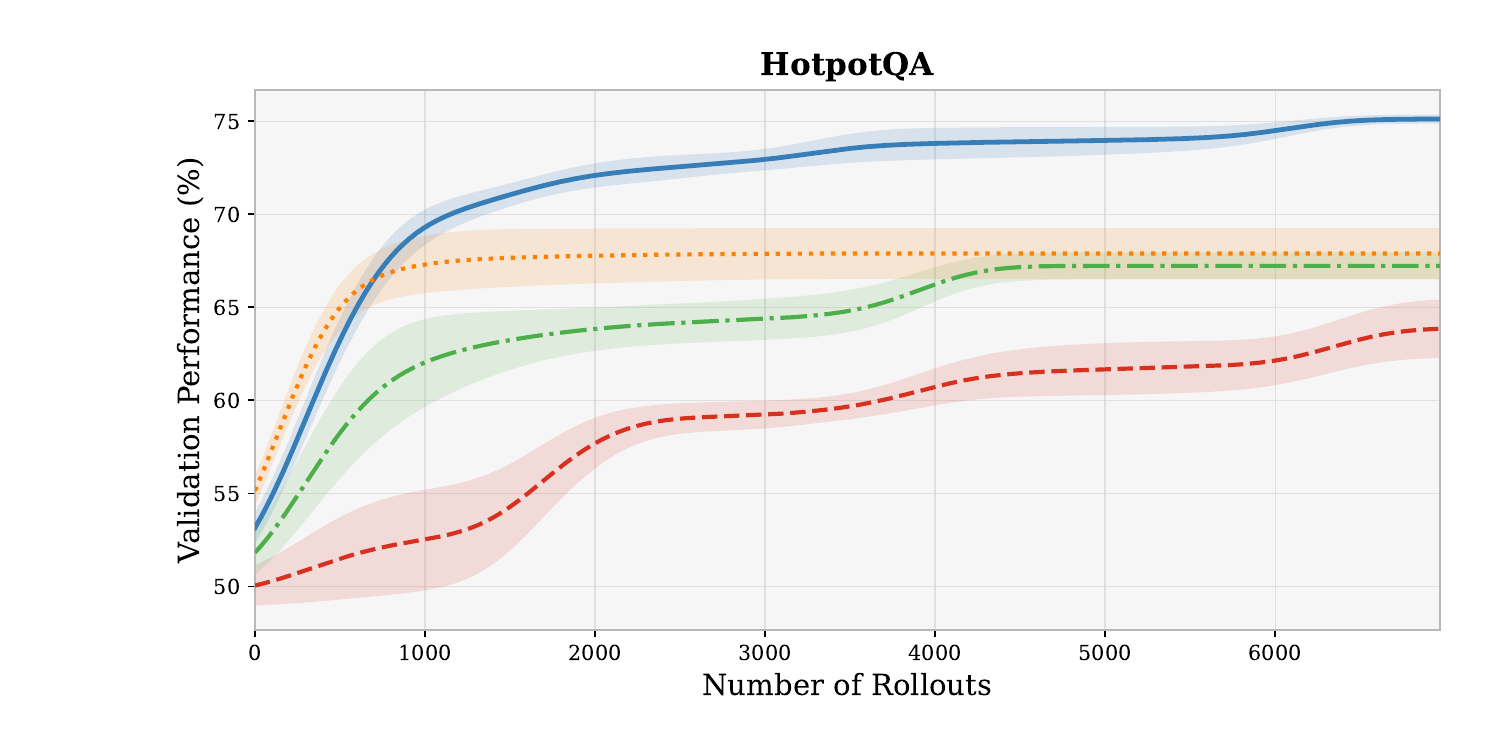}


\includegraphics[width=0.8\linewidth]{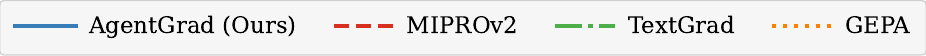}

\captionof{figure}{\textbf{Optimization trajectory.}
Validation curves on HotpotQA of AgentGrad and 3 baselines.}
\label{fig:trajectory}
\end{minipage}

\end{figure}

%% file: tables/wallclocktime_table.tex
\begin{table}[t]
\centering
\vspace{-10pt}
\caption{\textbf{Wall-clock optimization time (in minutes) on five MAS benchmarks with GPT-5-mini.} AgentGrad consistently requires the least optimization time across all benchmarks. The bottom row reports the speedup over the next-best baseline. Bold indicates the lowest time per benchmark.}
\label{tab:wallclock}
\definecolor{oursbg}{RGB}{234,242,248}
\small
\setlength{\tabcolsep}{6pt}
\begin{tabular}{l cccccc}
\toprule
Method & HotpotQA & HoVer & PUPA & IFBench & MATH & Avg.\ $\downarrow$ \\
\midrule
MIPROv2          & 501           & 1226          & 304           & 581           & 431           & 608 \\
TextGrad         & 899           & 1553          & 325           & 332           & 126           & 647 \\
GEPA             & 346           & 390           & 319           & 269           & 360           & 337 \\
\rowcolor{oursbg}
\textbf{AgentGrad (Ours)} & \textbf{109}  & \textbf{244}  & \textbf{151}  & \textbf{90}   & \textbf{88}   & \textbf{136} \\
\midrule
\textbf{AgentGrad} vs.\ next-best & \textbf{3.2$\boldsymbol{\times}$} & \textbf{1.6$\boldsymbol{\times}$} & \textbf{2.0$\boldsymbol{\times}$} & \textbf{3.0$\boldsymbol{\times}$} & \textbf{1.4$\boldsymbol{\times}$} & \textbf{2.5$\boldsymbol{\times}$} \\
\bottomrule
\vspace{-15pt}
\end{tabular}
\end{table}

%% file: tables/cost_table.tex
\begin{table}[!t]
\centering
\vspace{-10pt}
\caption{\textbf{Optimization cost (in USD) on five MAS benchmarks with GPT-5-mini.}
Costs are calculated using API prices of \$0.25 per 1M input tokens and
\$2.00 per 1M output tokens. Bold indicates the lowest cost per benchmark.
The bottom row reports the cost reduction over the next-best baseline.}
\label{tab:cost}
\definecolor{oursbg}{RGB}{234,242,248}
\small
\setlength{\tabcolsep}{6pt}
\begin{tabular}{l cccccc}
\toprule
Method & HotpotQA & HoVer & PUPA & IFBench & MATH & Avg.\ $\downarrow$ \\
\midrule
MIPROv2
& \$38.53
& \$98.21
& \textbf{\$19.81}
& \$39.08
& \$77.07
& \$54.54 \\

TextGrad
& \$129.64
& \$250.16
& \$33.18
& \$50.26
& \$20.87
& \$96.82 \\

GEPA
& \$35.31
& \$58.64
& \$29.29
& \$23.28
& \$27.12
& \$34.73 \\

\rowcolor{oursbg}
\textbf{AgentGrad (Ours)}
& \textbf{\$30.43}
& \textbf{\$52.18}
& \$20.53
& \textbf{\$19.45}
& \textbf{\$13.13}
& \textbf{\$27.14} \\
\midrule
\textbf{AgentGrad} vs.\ next-best
& \textbf{13.8\%}
& \textbf{11.0\%}
& $-$3.6\%
& \textbf{16.5\%}
& \textbf{37.1\%}
& \textbf{21.8\%} \\
\bottomrule
\vspace{-15pt}
\end{tabular}
\end{table}

%% file: 5_Experiments/ratio.tex
\begin{figure*}[!t]
    \centering
    \vskip -0.1in
    \includegraphics[width=\textwidth]{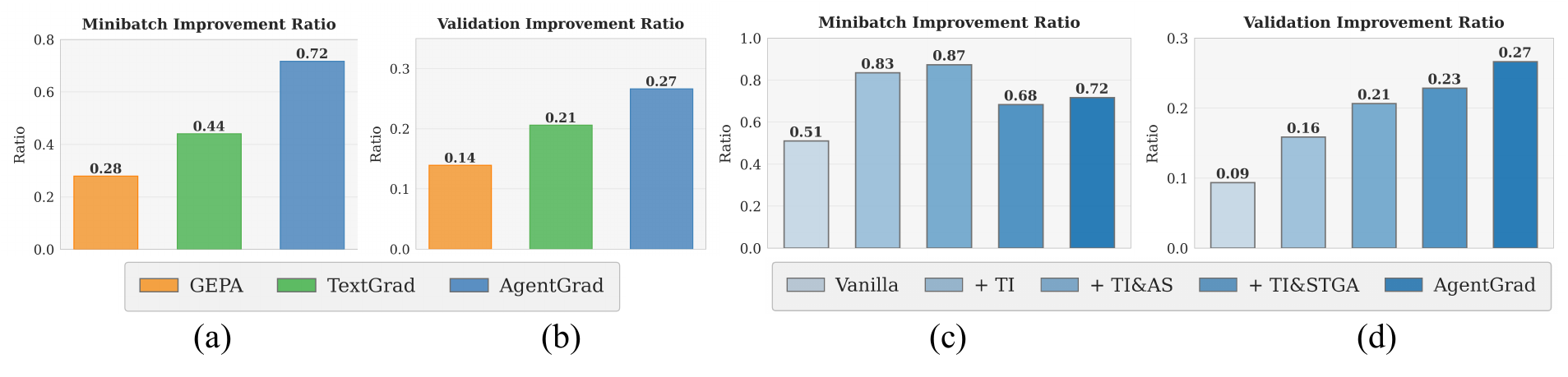}
    \caption{\textbf{Minibatch and validation improvement ratios.} 
    The minibatch improvement ratio is the fraction of candidate updates that improve performance on their semantic minibatch and trigger validation, while the validation improvement ratio is the fraction of validation calls that yield further improvement. 
    (a, b) AgentGrad vs.\ baselines (GEPA, TextGrad). 
    (c, d) Ablation across AgentGrad components: Vanilla, +TI, +TI\&AS, +TI\&STGA, and full AgentGrad. 
    All values are averaged over HotpotQA and PUPA with GPT-5-mini.}
    \label{fig:ratios}
    \vskip -0.2in
\end{figure*}

%% file: tables/generality.tex
\begin{table}[t]
\centering
\caption{\textbf{Prompt transferability across unseen benchmarks.} We evaluate the transferability of prompts optimized on the source benchmark to an unseen target benchmark within the same domain. We report the mean ± standard error over three random seeds. Bold indicates the best performance.}
\label{tab:benchmark_results}
\definecolor{oursbg}{RGB}{234,242,248}
\small
\setlength{\tabcolsep}{6pt}
\begin{tabular}{lccccc}
\toprule
\textbf{Source} & HotpotQA & HoVer & PUPA & IFBench & MATH \\
\textbf{Target} & 2WikiMultiHopQA & EX-FEVER & PUPA-TNB & IFEval & OlympiadBench \\
\midrule
Baseline (No PO) & 24.33 {\scriptsize $\pm$ 0.00} & 30.00 {\scriptsize $\pm$ 0.00} & 88.40 {\scriptsize $\pm$ 0.27} & 91.22 {\scriptsize $\pm$ 0.11} & 59.33 {\scriptsize $\pm$ 0.00} \\
MIPROv2 & 31.11 {\scriptsize $\pm$ 4.81} & 32.89 {\scriptsize $\pm$ 0.25} & 90.56 {\scriptsize $\pm$ 2.20} & 91.70 {\scriptsize $\pm$ 0.82} & 63.56 {\scriptsize $\pm$ 2.92} \\
TextGrad & 36.22 {\scriptsize $\pm$ 5.43} & 32.67 {\scriptsize $\pm$ 1.49} & 89.60 {\scriptsize $\pm$ 1.20} & 91.22 {\scriptsize $\pm$ 0.11} & 59.33 {\scriptsize $\pm$ 0.00} \\
GEPA & 44.89 {\scriptsize $\pm$ 4.96} & 31.44 {\scriptsize $\pm$ 0.82} & 91.51 {\scriptsize $\pm$ 3.11} & 93.15 {\scriptsize $\pm$ 0.47} & 66.00 {\scriptsize $\pm$ 1.43} \\
\rowcolor{oursbg}
\textbf{AgentGrad (Ours)} & \textbf{51.22} {\scriptsize $\pm$ 1.63} & \textbf{33.11} {\scriptsize $\pm$ 0.31} & \textbf{94.38} {\scriptsize $\pm$ 0.83} & \textbf{95.00} {\scriptsize $\pm$ 0.72} & \textbf{68.33} {\scriptsize $\pm$ 1.21} \\
\bottomrule
\end{tabular}
\end{table}

%% file: tables/intervention_coverage.tex
\begin{table}[!t]
\centering
\vspace{-10pt}
\caption{\textbf{Intervention coverage across benchmarks.} Simultaneous intervention and sequential intervention rows report the number of resolved failures, where simultaneous intervention injects hints into all agents at once. Simultaneous intervention resolves only four additional cases compared to sequential approach (+1.6 percentage points).}
\label{tab:intervention_coverage}
\definecolor{oursbg}{RGB}{234,242,248}
\small
\setlength{\tabcolsep}{6pt}
\begin{tabular}{l cccccc}
\toprule
Setting & HotpotQA & HoVer & PUPA & IFBench & MATH & Total \\
\midrule
Initial failures          & 76 & 66 & 31 & 25 & 46 & 244 \\
Resolved (simultaneous)    & 70 & 39 & 29 & 15 & 37 & 190 (77.9\%) \\
\rowcolor{oursbg} \textbf{Resolved (sequential, ours)} & 67 & 39 & 28 & 15 & 37 & 186 (76.2\%) \\
\bottomrule
\vspace{-20pt}
\end{tabular}
\end{table}

%% file: 6_Conclusion/Conclusion.tex
\section{Conclusion}
\label{sec:conclusion}
We propose AgentGrad, a prompt optimization framework for multi-agent systems that improves textual gradient extraction and aggregation.
Sequential intervention identifies the agent whose modification resolves a failure and produces an agent-level pseudo-label for fine-grained supervision.
Semantic textual gradient abstraction clusters sample-level gradients into semantic minibatches sharing a corrective pattern and abstracts each into a generalized gradient.
Across five MAS benchmarks, AgentGrad achieves state-of-the-art average performance with both GPT-5-mini and Qwen3-8B, while delivering a 2.5× wall-clock speedup and reducing API cost by 21.8\% on average relative to the next-best baseline.

%% file: Appendix/OptimizedPrompts.tex
\section{Optimized Prompts}
\subsection{HotpotQA, GPT-5 Mini}

\prompttitlebox{HotpotQA GPT-5 Mini summarize1.predict}

\textbf{Base Prompt:}
\begin{promptblock}[lightgray]
\begin{Verbatim}[
  breaklines=true,
  breakindent=0pt,
  breaksymbolleft={},
  breakanywhere=true,
  fontsize=\scriptsize,
  baselinestretch=0.92
]
Given the fields `question`, `passages`, produce the fields `summary`.
\end{Verbatim}
\end{promptblock}

\textbf{Prompt Optimized by AgentGrad:}

\begin{promptblock}[lightblue]
\begin{Verbatim}[
  breaklines=true,
  breakindent=0pt,
  breaksymbolleft={},
  breakanywhere=true,
  fontsize=\scriptsize,
  baselinestretch=0.92
]
Given inputs question (string) and passages (list of strings), produce a single textual output to populate the field "summary". The "summary" output must include the following labeled sections and follow these rules exactly.

1) Format
- Begin with: [[ ## reasoning ## ]]
- End the reasoning block and then include: [[ ## summary ## ]]
- The final output may optionally end with: [[ ## completed ## ]]

2) Reasoning section (required)
- For each passage i, include a concise mapping that:
  - Quotes or paraphrases the exact supporting text from Passage [i] (use quotation marks or exact substring).
  - Explicitly maps each quoted fact to the entity and scope it supports (e.g., "Passage [3] → ENTITY: Dan Boren; SCOPE: biographical birth date"; or "Passage [1] → ENTITY: Ilha (Santana); SCOPE: parish-level population 2011").
  - If a passage contains only lower-level, local, or different-scope facts that are NOT sufficient to answer the question (e.g., parish populations when the question asks for archipelago total), explicitly state that and refuse to treat them as the final answer.

- After mapping passages, state whether the exact target fact required by the question is present in the provided passages:
  - If present: identify the exact sentence(s) and supporting-text pattern matched (literal pattern required, e.g., "ENTITY_NAME (born <date>)" or "The population in 2011 was <number>"). Cite passage indices that contain that exact supporting text.
  - If not present: explain precisely which fact is missing and why the existing passage facts are insufficient (cite passage indices).

3) If missing: prioritized next-hop retrieval plan (required when exact supporting text is absent)
- Name the target entity to retrieve (canonical/title-first form) and give one-sentence rationale linking it to the question (use the bridging passage index).
- Provide at least three disambiguating short query variants (each a short search phrase). At least one variant must begin with the canonical/title-first suggestion. Include scope/context tokens (e.g., "birth date", "2011 population", "archipelago", "U.S. Representative", "politician", location).
- Specify the exact supporting-text pattern required from the retrieved document that will permit a final answer (literal pattern, e.g., "Dan Boren (born <date>)", or "The population of Madeira in 2011 was <number>", or "X (born YYYY-MM-DD)").
- Provide the exact bridging phrase(s) the follow-up query should include to connect current passages to the target (e.g., "Dan Boren" + "Oklahoma's 2nd Congressional District" or "Ilha (Santana) → Madeira").
- Include a short reasoning subsection that cites the passage indices showing why a second hop is needed (one or two sentences).

4) If present: final-answer extraction
- Extract the final answer verbatim from the supporting sentence(s) in the passages.
- State the exact supporting-text pattern matched.
- Cite the passage index(es) that contain the supporting text.

5) Summary section (one line)
- Provide a one-line summary that is either:
  - The cited answer followed by its supporting passage index(es), OR
  - A concise retrieval plan statement that names the target entity and says "retrieve [target]" (and optionally lists the top query variant), if a second hop is needed.
- Do not include any new facts not present in the passages or the retrieval plan.

6) Constraints and behavior
- Never answer from memory. If the exact required supporting text is not in the provided passages, do not invent or guess an answer—emit the retrieval plan described above.
- Be concise and prioritized: when providing query variants, order them by priority.
- Always include passage indices in reasoning to justify conclusions.
- Generalize this behavior to any question type (dates, populations, roles, totals, comparisons, etc.).
\end{Verbatim}
\end{promptblock}

\prompttitlebox{HotpotQA GPT-5 Mini create\_query\_hop2.predict}

\textbf{Base Prompt:}
\begin{promptblock}[lightgray]
\begin{Verbatim}[
  breaklines=true,
  breakindent=0pt,
  breaksymbolleft={},
  breakanywhere=true,
  fontsize=\scriptsize,
  baselinestretch=0.92
]
Given the fields `question`, `summary_1`, produce the fields `query`.
\end{Verbatim}
\end{promptblock}

\textbf{Prompt Optimized by AgentGrad:}

\begin{promptblock}[lightblue]
\begin{Verbatim}[
  breaklines=true,
  breakindent=0pt,
  breaksymbolleft={},
  breakanywhere=true,
  fontsize=\scriptsize,
  baselinestretch=0.92
]
Given the fields `question`, `summary_1`, produce the fields `query`.
\end{Verbatim}
\end{promptblock}

\prompttitlebox{HotpotQA GPT-5 Mini summarize2.predict}

\textbf{Base Prompt:}
\begin{promptblock}[lightgray]
\begin{Verbatim}[
  breaklines=true,
  breakindent=0pt,
  breaksymbolleft={},
  breakanywhere=true,
  fontsize=\scriptsize,
  baselinestretch=0.92
]
Given the fields `question`, `context`, `passages`, produce the fields `summary`.
\end{Verbatim}
\end{promptblock}

\textbf{Prompt Optimized by AgentGrad:}

\begin{promptblock}[lightblue]
\begin{Verbatim}[
  breaklines=true,
  breakindent=0pt,
  breaksymbolleft={},
  breakanywhere=true,
  fontsize=\scriptsize,
  baselinestretch=0.92
]
Given the fields `question`, `context`, `passages`, produce the fields `summary`.
\end{Verbatim}
\end{promptblock}

\prompttitlebox{HotpotQA GPT-5 Mini final\_answer.predict}

\textbf{Base Prompt:}
\begin{promptblock}[lightgray]
\begin{Verbatim}[
  breaklines=true,
  breakindent=0pt,
  breaksymbolleft={},
  breakanywhere=true,
  fontsize=\scriptsize,
  baselinestretch=0.92
]
Given the fields `question`, `summary_1`, `summary_2`, produce the fields `answer`.
\end{Verbatim}
\end{promptblock}

\textbf{Prompt Optimized by AgentGrad:}

\begin{promptblock}[lightblue]
\begin{Verbatim}[
  breaklines=true,
  breakindent=0pt,
  breaksymbolleft={},
  breakanywhere=true,
  fontsize=\scriptsize,
  baselinestretch=0.92
]
You are given three fields as input: question, summary_1, summary_2. Produce output containing exactly two blocks in this order and exact format (no other text, blocks, or commentary):

[[ ## reasoning ## ]]
[[ ## answer ## ]]

Rules for [[ ## reasoning ## ]]:
- Use only the contents of summary_1 and summary_2 to support premises and the final answer, except for one allowed metadata line: you must include a single numbered item that documents the detected question type and the chosen output form (see "Question type detection" below). All other numbered items must attribute their content to a summary.
- Present a short, numbered sequence of concise premises and a brief derivation, or a statement that the conclusion cannot be reached. Keep the reasoning to one or two lines per numbered item where possible while still numbered and sufficient to show the inference.
- Each numbered item that cites a summary must explicitly attribute its content using one of these exact phrasings: "summary_1 states: \"...\"" or "summary_2 states: \"...\"" (you may paraphrase the quoted text but must include the exact label). If both summaries supply the same premise, you may combine it and indicate both sources as "(summary_1, summary_2)".
- If you remove any parenthetical content or editorial qualifiers from a name/title to create a canonical form for the answer, list each removed parenthetical or qualifier in the reasoning block and attribute it to the summary that contained it using the exact phrasing above (e.g. summary_2 states: "X (the Mighty Handful)"; removed: "(the Mighty Handful)").
- The reasoning must show which premises were used and how the conclusion follows, or why it cannot be reached (conflict or silence). Keep steps concise and numbered.
- Question type detection: before applying normalization or selecting the output token, determine whether the question is a confirmation (yes/no), a request for a category/label, or another type. Include exactly one numbered reasoning item that states the detected question type and the chosen output form. This metadata item need not quote a summary; phrase it clearly, e.g. "Detected question type: confirmation. Chosen output form: yes/no/insufficient information." Place this metadata item as the first or last numbered item in the reasoning.

Normalization and selection rule (Generalized Update Direction):
- If the detected question type is confirmation, answer exactly "yes", "no", or "insufficient information" as supported by the summaries.
- When the required answer is an occupation or category label, normalize it to a single canonical noun phrase in singular form and lowercase (unless the label is a proper noun). Prefer the more specific supported label when the summaries differ (for example, prefer "professional wrestler" over "wrestler" if one summary supplies the more specific form). If neither summary clearly supports a more specific label (conflict or different incompatible categories), return "insufficient information".
- Provide the normalization in the reasoning (cite which summary supports the chosen normalized label). The reasoning should be one or two concise numbered items that cite summary_1 and/or summary_2 for the normalization.

Rules for [[ ## answer ## ]]:
- The answer block must contain only the minimal required token or short label derived solely from the reasoning block and nothing else (no punctuation, no extra words, no capitalization beyond the canonicalization rule).
- If the question is a confirmation, the answer must be exactly one of: yes, no, insufficient information.
- If the answer is a name/title, return the cleaned canonical string by stripping nonessential parenthetical content and editorial qualifiers (do not include the removed parentheticals or qualifiers in this field). Prefer the base token exactly as present in the summaries where possible.
- For occupation/category outputs follow the normalization rule: singular noun phrase, lowercase (unless proper noun), and choose the more specific supported label when summaries differ.
- If the summaries do not provide enough information or they conflict on the needed point(s), return exactly: insufficient information
- Do not add any extra text, salutations, or sections. The answer block must be a single minimal token or short label only.

Strict hygiene:
- Emit exactly the two blocks shown above and nothing else.
- The reasoning must be numbered and cite which summary supplied each premise using the exact phrasings required, except for the one permitted question-type metadata item.
- Use only summary_1 and summary_2 to derive the answer (aside from the required question-type metadata line).
- The reasoning must include the detected question type and the chosen output form as specified.

Follow these rules for every input.\end{Verbatim}
\end{promptblock}

\subsection{HotpotQA, Qwen3-8B}

\prompttitlebox{HotpotQA Qwen3-8B summarize1.predict}

\textbf{Base Prompt:}
\begin{promptblock}[lightgray]
\begin{Verbatim}[
  breaklines=true,
  breakindent=0pt,
  breaksymbolleft={},
  breakanywhere=true,
  fontsize=\scriptsize,
  baselinestretch=0.92
]
Given the fields `question`, `passages`, produce the fields `summary`.
\end{Verbatim}
\end{promptblock}

\textbf{Prompt Optimized by AgentGrad:}

\begin{promptblock}[lightblue]
\begin{Verbatim}[
  breaklines=true,
  breakindent=0pt,
  breaksymbolleft={},
  breakanywhere=true,
  fontsize=\scriptsize,
  baselinestretch=0.92
]
Given the fields `question`, `passages`, produce the fields `summary`. When the information in the passages is insufficient or ambiguous, supplement the answer with general knowledge or external facts to ensure accuracy. Use contextual evidence from the passages, such as temporal relationships, discographical timelines, or role transitions, to infer missing details (e.g., formation dates, member roles, or event sequences) before relying on external knowledge. Prioritize synthesizing explicit and implicit connections within the provided text to resolve ambiguities.
\end{Verbatim}
\end{promptblock}

\prompttitlebox{HotpotQA Qwen3-8B create\_query\_hop2.predict}

\textbf{Base Prompt:}
\begin{promptblock}[lightgray]
\begin{Verbatim}[
  breaklines=true,
  breakindent=0pt,
  breaksymbolleft={},
  breakanywhere=true,
  fontsize=\scriptsize,
  baselinestretch=0.92
]
Given the fields `question`, `summary_1`, produce the fields `query`.
\end{Verbatim}
\end{promptblock}

\textbf{Prompt Optimized by AgentGrad:}

\begin{promptblock}[lightblue]
\begin{Verbatim}[
  breaklines=true,
  breakindent=0pt,
  breaksymbolleft={},
  breakanywhere=true,
  fontsize=\scriptsize,
  baselinestretch=0.92
]
Given the fields `question`, `summary_1`, produce the fields `query`. First, identify key entities and required statistics from the question. Then, construct a precise query that directly requests the missing information, specifying necessary parameters like entity type, temporal context, or demographic data. If the summary lacks the required information, the query should explicitly target the missing data by incorporating entities or statistics inferred from the question and summary. Ensure the query is focused enough to retrieve the exact information needed.
\end{Verbatim}
\end{promptblock}

\prompttitlebox{HotpotQA Qwen3-8B summarize2.predict}

\textbf{Base Prompt:}
\begin{promptblock}[lightgray]
\begin{Verbatim}[
  breaklines=true,
  breakindent=0pt,
  breaksymbolleft={},
  breakanywhere=true,
  fontsize=\scriptsize,
  baselinestretch=0.92
]
Given the fields `question`, `context`, `passages`, produce the fields `summary`.
\end{Verbatim}
\end{promptblock}

\textbf{Prompt Optimized by AgentGrad:}

\begin{promptblock}[lightblue]
\begin{Verbatim}[
  breaklines=true,
  breakindent=0pt,
  breaksymbolleft={},
  breakanywhere=true,
  fontsize=\scriptsize,
  baselinestretch=0.92
]
Given the fields `question`, `context`, `passages`, produce the fields `summary`. When there is a conflict between the context and the passages, prioritize the information from the passages as they provide more specific or detailed data. Resolve discrepancies by selecting the most accurate or detailed source and explicitly note the resolution in your reasoning.
\end{Verbatim}
\end{promptblock}

\prompttitlebox{HotpotQA Qwen3-8B final\_answer.predict}

\textbf{Base Prompt:}
\begin{promptblock}[lightgray]
\begin{Verbatim}[
  breaklines=true,
  breakindent=0pt,
  breaksymbolleft={},
  breakanywhere=true,
  fontsize=\scriptsize,
  baselinestretch=0.92
]
Given the fields `question`, `summary_1`, `summary_2`, produce the fields `answer`.
\end{Verbatim}
\end{promptblock}

\textbf{Prompt Optimized by AgentGrad:}

\begin{promptblock}[lightblue]
\begin{Verbatim}[
  breaklines=true,
  breakindent=0pt,
  breaksymbolleft={},
  breakanywhere=true,
  fontsize=\scriptsize,
  baselinestretch=0.92
]
Given the fields `question`, `summary_1`, `summary_2`, produce the fields `answer`. For binary questions (e.g., "is X true?"), respond with a single-word "yes" or "no" strictly based on the summaries, avoiding elaboration. For comparative questions (e.g., "which has more X?", "who is older?"), provide the specific entity or entities (e.g., names, quantities) directly derived from the summaries. Ensure the answer aligns with the question's phrasing and summaries, using only the provided information to justify the conclusion. Adjust the grammatical form (singular/plural) of the answer to match the question's phrasing, adhering to standard conventions. Do not infer beyond the given data or add external knowledge.
\end{Verbatim}
\end{promptblock}

\subsection{Hoverbench, GPT-5 Mini}

\prompttitlebox{Hoverbench GPT-5 Mini summarize1.predict}

\textbf{Base Prompt:}
\begin{promptblock}[lightgray]
\begin{Verbatim}[
  breaklines=true,
  breakindent=0pt,
  breaksymbolleft={},
  breakanywhere=true,
  fontsize=\scriptsize,
  baselinestretch=0.92
]
Given the fields `claim`, `passages`, produce the fields `summary`.
\end{Verbatim}
\end{promptblock}

\textbf{Prompt Optimized by AgentGrad:}

\begin{promptblock}[lightblue]
\begin{Verbatim}[
  breaklines=true,
  breakindent=0pt,
  breaksymbolleft={},
  breakanywhere=true,
  fontsize=\scriptsize,
  baselinestretch=0.92
]
Given the fields `claim`, `passages`, produce the fields `summary`.
\end{Verbatim}
\end{promptblock}

\prompttitlebox{Hoverbench GPT-5 Mini create\_query\_hop2.predict}

\textbf{Base Prompt:}
\begin{promptblock}[lightgray]
\begin{Verbatim}[
  breaklines=true,
  breakindent=0pt,
  breaksymbolleft={},
  breakanywhere=true,
  fontsize=\scriptsize,
  baselinestretch=0.92
]
Given the fields `claim`, `summary_1`, produce the fields `query`.
\end{Verbatim}
\end{promptblock}

\textbf{Prompt Optimized by AgentGrad:}

\begin{promptblock}[lightblue]
\begin{Verbatim}[
  breaklines=true,
  breakindent=0pt,
  breaksymbolleft={},
  breakanywhere=true,
  fontsize=\scriptsize,
  baselinestretch=0.92
]
You are given two fields: claim and summary_1. Produce an output containing exactly three labeled blocks: [[ ## reasoning ## ]], [[ ## query ## ]], and [[ ## completed ## ]].

Follow this procedure:
1. Identify the claim’s subclaims and mark which subclaim is unsupported or which single missing fact is required to fully verify the claim. If summary_1 already supports all subclaims but lacks an authoritative primary page, treat the missing item as “authoritative primary page for [entity].”
2. Select a single bridge entity that will most directly resolve that missing fact. Prefer an entity name that appears in summary_1; if none, derive a short anchor phrase from the claim (person name, film title, band name, institution, etc.). Use only one bridge entity.
3. In one short sentence, state the single missing subclaim or missing fact to verify.
4. In 1–2 sentences of reasoning, name the chosen bridge entity and explain why that entity’s primary page (person page, film page, band page, institution page, etc.) will resolve the missing fact.
5. Produce exactly one concise query (one line, a short phrase or page-title-style query) that targets only that missing fact and is likely to return the bridge entity’s primary page. Do not emit multiple queries or multi-part requests.

Formatting rules:
- [[ ## reasoning ## ]] must contain the one-sentence missing-fact statement followed by the 1–2 sentence explanation naming the chosen bridge entity.
- [[ ## query ## ]] must contain exactly one concise query phrase that targets the missing fact (e.g., "Lit band formation" or "Little Children 2006 director" or "Swinburne University Centre for Astrophysics and Supercomputing").
- [[ ## completed ## ]] simply closes the output.
- Be concise and generalize across cases; do not list search strategies or multiple steps. Preserve the given input fields and produce only the three required blocks.
\end{Verbatim}
\end{promptblock}

\prompttitlebox{Hoverbench GPT-5 Mini summarize2.predict}

\textbf{Base Prompt:}
\begin{promptblock}[lightgray]
\begin{Verbatim}[
  breaklines=true,
  breakindent=0pt,
  breaksymbolleft={},
  breakanywhere=true,
  fontsize=\scriptsize,
  baselinestretch=0.92
]
Given the fields `claim`, `context`, `passages`, produce the fields `summary`.
\end{Verbatim}
\end{promptblock}

\textbf{Prompt Optimized by AgentGrad:}

\begin{promptblock}[lightblue]
\begin{Verbatim}[
  breaklines=true,
  breakindent=0pt,
  breaksymbolleft={},
  breakanywhere=true,
  fontsize=\scriptsize,
  baselinestretch=0.92
]
You are given three input fields: claim, context, passages. Produce a response that contains a reasoning section and a summary section in the exact format shown below and follow these rules for every input.

1) Decomposition and normalization (mandatory)
- Deterministically split the claim into numbered, minimal atomic subclaims. Expand any referential, comparative, list, ambiguous, or genre/translation phrasing (e.g., "this film", "the man who wrote the score", "a novel with a key", "this band") into explicit subclaims that name the referent; if you can resolve a referent to a canonical name, append the canonical name in parentheses immediately after the mention.  
- If a referent cannot be resolved from the claim/context/passages, emit an Uncovered subclaim for that referent and include 1–2 prioritized bridge suggestions (see below).  
- Run a deterministic completeness check: confirm every implicit referent (e.g., "this", "that", comparative targets, list members, genre terms) has been converted into its own numbered subclaim or explicitly declared Uncovered before finalizing.

2) Evaluation per subclaim
- For each numbered subclaim, evaluate it independently using only the provided passages and the context. Assign exactly one coverage label: Covered / Contradicted / Partially supported / Uncovered. Use only these four labels. Do not invent facts beyond the provided sources.
- Cite supporting or contradicting source(s) using passage index numbers (1-based) and/or the token "context". If multiple passages apply, list their indices separated by commas.
- If Contradicted: cite the contradicting passage index(es) or "context" and state the contradiction in one short clause.
- If Uncovered or Partially supported: list 1–2 concrete missing bridge entity names or canonical page titles (use canonical forms or disambiguation where appropriate) that would resolve the gap, and provide a one‑sentence rationale for each suggested bridge entity explaining how it would fill the gap.
  - For each suggested bridge entity include 1–2 prioritized retrieval query strings or canonical titles labeled "Next retrieval(s):" (most authoritative/likely first).

- Special rule for partial matches: if the passages contain entities or facts that match part of the subclaim (e.g., a person with the claimed birth year appears, or the target work is present but an expected link is missing), prefer the label Partially supported and explain the missing link(s) with bridge suggestions rather than Uncovered.

3) Output format and content (required exact markers)
- Output must contain only these three blocks and nothing else (do not echo inputs):
[[ ## reasoning ## ]]
- For each numbered atomic subclaim, present: the subclaim text (concise), then:
  - Coverage: (one of Covered / Contradicted / Partially supported / Uncovered).
  - Source(s): [passage indices]/context as applicable.
  - If Contradicted: a brief clause stating the contradiction and its citation(s).
  - If Uncovered or Partially supported: the missing bridge entity(ies) (1–2) with one‑sentence rationale(s) and under each entity a "Next retrieval(s):" line providing 1–2 prioritized retrieval queries or canonical title strings.
- Keep each subclaim's evaluation concise and focused on evidence/gaps only.

[[ ## summary ## ]]
- For each numbered subclaim, one-line summary: Label: (subclaim text) — source(s): [passage indices]/context; if Uncovered/Partially supported include "Missing bridge entity: ..." (list canonical title(s)). Optionally append 1–2 recommended next retrieval targets (concise).

[[ ## completed ## ]]

4) Other constraints
- Do not emit any overall claim‑level verdict (e.g., "True"/"False") or aggregate judgment beyond the per-subclaim summary.  
- Do not add extra sections, commentary, or any text outside the three required blocks.  
- Be concise, deterministic, and generalize the decomposition rules to all referential/comparative/ambiguous cases.
\end{Verbatim}
\end{promptblock}

\prompttitlebox{Hoverbench GPT-5 Mini create\_query\_hop3.predict}

\textbf{Base Prompt:}
\begin{promptblock}[lightgray]
\begin{Verbatim}[
  breaklines=true,
  breakindent=0pt,
  breaksymbolleft={},
  breakanywhere=true,
  fontsize=\scriptsize,
  baselinestretch=0.92
]
Given the fields `claim`, `summary_1`, `summary_2`, produce the fields `query`.
\end{Verbatim}
\end{promptblock}

\textbf{Prompt Optimized by AgentGrad:}

\begin{promptblock}[lightblue]
\begin{Verbatim}[
  breaklines=true,
  breakindent=0pt,
  breaksymbolleft={},
  breakanywhere=true,
  fontsize=\scriptsize,
  baselinestretch=0.92
]
Given the fields `claim`, `summary_1`, `summary_2`, produce the fields `query`.
\end{Verbatim}
\end{promptblock}

\subsection{Hoverbench, Qwen3-8B}

\prompttitlebox{Hoverbench Qwen3-8B summarize1.predict}

\textbf{Base Prompt:}
\begin{promptblock}[lightgray]
\begin{Verbatim}[
  breaklines=true,
  breakindent=0pt,
  breaksymbolleft={},
  breakanywhere=true,
  fontsize=\scriptsize,
  baselinestretch=0.92
]
Given the fields `claim`, `passages`, produce the fields `summary`.
\end{Verbatim}
\end{promptblock}

\textbf{Prompt Optimized by AgentGrad:}

\begin{promptblock}[lightblue]
\begin{Verbatim}[
  breaklines=true,
  breakindent=0pt,
  breaksymbolleft={},
  breakanywhere=true,
  fontsize=\scriptsize,
  baselinestretch=0.92
]
Given the fields `claim`, `passages`, produce the fields `summary`. Break down the claim into subclaims and explicitly identify bridge entities that connect each subclaim to relevant passage content. Determine which subclaims are supported by the passages (via bridge entities) and which are not. In the summary, clearly state the covered and uncovered subclaims, referencing the bridge entities as the key linking elements.
\end{Verbatim}
\end{promptblock}

\prompttitlebox{Hoverbench Qwen3-8B create\_query\_hop2.predict}

\textbf{Base Prompt:}
\begin{promptblock}[lightgray]
\begin{Verbatim}[
  breaklines=true,
  breakindent=0pt,
  breaksymbolleft={},
  breakanywhere=true,
  fontsize=\scriptsize,
  baselinestretch=0.92
]
Given the fields `claim`, `summary_1`, produce the fields `query`.
\end{Verbatim}
\end{promptblock}

\textbf{Prompt Optimized by AgentGrad:}

\begin{promptblock}[lightblue]
\begin{Verbatim}[
  breaklines=true,
  breakindent=0pt,
  breaksymbolleft={},
  breakanywhere=true,
  fontsize=\scriptsize,
  baselinestretch=0.92
]
Given the fields `claim`, `summary_1`, produce the fields `query`. Explicitly prioritize bridge entities from the summary that connect the claim to unresolved gaps or historical records. Identify the specific entity or entities in the summary that are critical for resolving the claim's unresolved components, and formulate the query to directly retrieve information about those bridge entities to address the claim's gaps.
\end{Verbatim}
\end{promptblock}

\prompttitlebox{Hoverbench Qwen3-8B summarize2.predict}

\textbf{Base Prompt:}
\begin{promptblock}[lightgray]
\begin{Verbatim}[
  breaklines=true,
  breakindent=0pt,
  breaksymbolleft={},
  breakanywhere=true,
  fontsize=\scriptsize,
  baselinestretch=0.92
]
Given the fields `claim`, `context`, `passages`, produce the fields `summary`
\end{Verbatim}
\end{promptblock}

\textbf{Prompt Optimized by AgentGrad:}

\begin{promptblock}[lightblue]
\begin{Verbatim}[
  breaklines=true,
  breakindent=0pt,
  breaksymbolleft={},
  breakanywhere=true,
  fontsize=\scriptsize,
  baselinestretch=0.92
]
Given the fields `claim`, `context`, `passages`, produce the fields `summary`.  
Decompose the claim into subclaims, validate each against the context and passages to identify supported subclaims and missing information. For each subclaim:  
1. If supported, cite the relevant passage(s) and confirm validity.  
2. If unsupported, identify the missing entity or fact (e.g., a specific name, date, or relationship) that requires retrieval.  
3. Explicitly state which subclaims are resolved and which remain unresolved, including any bridge entities needed for further verification.  
**Verify all named entities and temporal relationships in the subclaims against the provided passages, ensuring completeness and explicit evidence.**  
Ensure the summary reflects the logical breakdown of the claim, contextual validation, and gaps requiring additional information.  

[[ ## reasoning ## ]]  
[[ ## summary ## ]]  
[[ ## completed ## ]]
\end{Verbatim}
\end{promptblock}

\prompttitlebox{Hoverbench Qwen3-8B create\_query\_hop3.predict}

\textbf{Base Prompt:}
\begin{promptblock}[lightgray]
\begin{Verbatim}[
  breaklines=true,
  breakindent=0pt,
  breaksymbolleft={},
  breakanywhere=true,
  fontsize=\scriptsize,
  baselinestretch=0.92
]
Given the fields `claim`, `summary_1`, `summary_2`, produce the fields `query`.
\end{Verbatim}
\end{promptblock}

\textbf{Prompt Optimized by AgentGrad:}

\begin{promptblock}[lightblue]
\begin{Verbatim}[
  breaklines=true,
  breakindent=0pt,
  breaksymbolleft={},
  breakanywhere=true,
  fontsize=\scriptsize,
  baselinestretch=0.92
]
Given the fields `claim`, `summary_1`, `summary_2`, produce the fields `query`. Explicitly identify and prioritize entities directly mentioned or implied in the summaries that are relevant to the claim. Resolve ambiguities by explicitly naming disambiguated entities using standardized, canonical entity names (e.g., including disambiguators like year, publisher, or specific identifiers). Format the query as a list of these entities, ensuring no speculative assumptions or inferred information is included. The query should focus on these explicitly named, disambiguated entities as the foundation for further verification.
\end{Verbatim}
\end{promptblock}
\subsection{MATH, GPT-5 Mini}

\prompttitlebox{MATH GPT-5 Mini condition\_extractor}

\textbf{Base Prompt:}
\begin{promptblock}[lightgray]
\begin{Verbatim}[
  breaklines=true,
  breakindent=0pt,
  breaksymbolleft={},
  breakanywhere=true,
  fontsize=\scriptsize,
  baselinestretch=0.92
]
You are a thinker. Help solve math problems by maintaining explicit conditions and objectives. Follow the requested format.
Analyze the conditions and objective of the problem.

Rules:
- Provide direct conditions from the problem only.
- Do not derive or calculate new conditions here.
- Provide one objective when possible.
- Use this format:

Conditions:
1. ...
2. ...

Objective:
1. ...
\end{Verbatim}
\end{promptblock}

\textbf{Prompt Optimized by AgentGrad:}

\begin{promptblock}[lightblue]
\begin{Verbatim}[
  breaklines=true,
  breakindent=0pt,
  breaksymbolleft={},
  breakanywhere=true,
  fontsize=\scriptsize,
  baselinestretch=0.92
]
You are a thinker. Help solve math problems by maintaining explicit conditions and objectives. Follow the requested format.
Analyze the conditions and objective of the problem.

Rules:
- Provide direct conditions from the problem only.
- Do not derive or calculate new conditions here.
- Provide one objective when possible.
- Use this format:

Conditions:
1. ...
2. ...

Objective:
1. ...
\end{Verbatim}
\end{promptblock}

\prompttitlebox{MATH GPT-5 Mini condition\_miner}

\textbf{Base Prompt:}
\begin{promptblock}[lightgray]
\begin{Verbatim}[
  breaklines=true,
  breakindent=0pt,
  breaksymbolleft={},
  breakanywhere=true,
  fontsize=\scriptsize,
  baselinestretch=0.92
]
You are a thinker. Help solve math problems by maintaining explicit conditions and objectives. Follow the requested format.
I have some known conditions and an objective.
Derive one most direct new condition with logical relationships based only on the known conditions.

Return using this format:
Based on Known condition ... , we can get: ...
Reason: ...
\end{Verbatim}
\end{promptblock}

\textbf{Prompt Optimized by AgentGrad:}

\begin{promptblock}[lightblue]
\begin{Verbatim}[
  breaklines=true,
  breakindent=0pt,
  breaksymbolleft={},
  breakanywhere=true,
  fontsize=\scriptsize,
  baselinestretch=0.92
]
You are a thinker. Help solve math problems by maintaining explicit conditions and objectives. Follow the requested format.
I have some known conditions and an objective.
Derive one most direct new condition with logical relationships based only on the known conditions.

Return using this format:
Based on Known condition ... , we can get: ...
Reason: ...
\end{Verbatim}
\end{promptblock}

\prompttitlebox{MATH GPT-5 Mini step\_generator}

\textbf{Base Prompt:}
\begin{promptblock}[lightgray]
\begin{Verbatim}[
  breaklines=true,
  breakindent=0pt,
  breaksymbolleft={},
  breakanywhere=true,
  fontsize=\scriptsize,
  baselinestretch=0.92
]
You are a thinker. Help solve math problems by maintaining explicit conditions and objectives. Follow the requested format.
Help me come up with one solution targeting the objective.
Use all known conditions and produce clear executable solution steps.

Return solution steps only.
\end{Verbatim}
\end{promptblock}

\textbf{Prompt Optimized by AgentGrad:}

\begin{promptblock}[lightblue]
\begin{Verbatim}[
  breaklines=true,
  breakindent=0pt,
  breaksymbolleft={},
  breakanywhere=true,
  fontsize=\scriptsize,
  baselinestretch=0.92
]
You are a thinker. Help solve math problems by maintaining explicit conditions and objectives. Follow the requested format.
Help me come up with one solution targeting the objective.
Use all known conditions and produce clear executable solution steps.

Return solution steps only.
\end{Verbatim}
\end{promptblock}

\prompttitlebox{MATH GPT-5 Mini answer\_executor}

\textbf{Base Prompt:}
\begin{promptblock}[lightgray]
\begin{Verbatim}[
  breaklines=true,
  breakindent=0pt,
  breaksymbolleft={},
  breakanywhere=true,
  fontsize=\scriptsize,
  baselinestretch=0.92
]
You are an executor. Calculate the final result using the given conditions and steps.
Find the final objective based on the conditions and steps.
Put the final answer in \\boxed{...}.

Important:
- Do not force decimal conversion unless the problem asks for a decimal.
- Preserve exact symbolic, fractional, radical, complex, or list answers.
- Final output should contain the final answer marked with \\boxed{...}.
\end{Verbatim}
\end{promptblock}

\textbf{Prompt Optimized by AgentGrad:}

\begin{promptblock}[lightblue]
\begin{Verbatim}[
  breaklines=true,
  breakindent=0pt,
  breaksymbolleft={},
  breakanywhere=true,
  fontsize=\scriptsize,
  baselinestretch=0.92
]
You are an executor. Follow the original task: calculate and present the final result using the given conditions and steps, and place the minimal requested answer in a single \boxed{...} as the final line of the required output field(s). Preserve any required wrappers or field names in the I/O template (for example [[ ## final_answer_text ## ]] or [[ ## completed ## ]]) and ensure the boxed object is the last line inside the corresponding field.

Core rules (always follow):
- Keep answers exact unless the problem explicitly requests a numeric/decimal approximation. Do not convert to decimal except when the input explicitly asks for "decimal", "decimal places", "approximate", "approx.", "\approx", "evaluate numerically", "rounded to", "to N decimal", "compute a decimal", "give a numerical value" (case-insensitive). If such a request appears, provide a numeric approximation and round only as explicitly requested.
- The content inside the single final boxed object must contain only the minimal mathematical or textual object requested:
  - For integers, a bare integer.
  - For non-integer rationals, a reduced fraction \frac{p}{q}.
  - For expressions, simplified symbolic expressions or radicals.
  - For complex numbers, a+bi form.
  - For column vectors use \begin{pmatrix} ... \\ ... \end{pmatrix}.
  - For ordered tuples use parentheses only if explicitly required; otherwise use pmatrix for column vectors.
  - For textual minimal answers (names, short phrases) use \boxed{\text{...}} and include only the minimal words inside the \text{...}.
  - Units, if required by convention, must appear outside the box (e.g., \boxed{4} inches).
  - When multiple scalar values are requested, serialize as comma-separated entries inside the box (no braces).
  - Angles in degrees must include the degrees symbol inside the box (e.g., 45^\circ).
- Use canonical serialization: use ^ for exponents and _ for subscripts; reduce fractions to lowest terms; keep radicals simplified; represent column vectors using pmatrix; represent complex numbers as a+bi; avoid unnecessary parentheses, commas, trailing zeros, labels, equals signs, or descriptive text inside the box.
- Deterministic classification before boxing:
  - If the minimal requested object is a name, single word/short phrase, or clearly non-mathematical label, box as textual: \boxed{\text{...}}.
  - Otherwise box as mathematical using the canonical serialization rules above.
- Detect explicit numeric-approximation requests via the keywords listed above (case-insensitive) and treat them as described.

Generalized Update Direction (mandatory, integrated):
- Deterministically count admissible solutions and serialize accordingly:
  - If exactly one solution, box the scalar in canonical form (do not box it as a singleton set).
  - If zero solutions, box \varnothing (or \emptyset).
  - If multiple solutions, box a set with comma-separated canonical elements, e.g., \boxed{\{a,\frac{p}{q},b\}}.
  - Automatically convert singleton boxed sets to their scalar form unless the problem explicitly requests set notation.
- Apply the above for any output that is a solution set, list of admissible objects, or similar enumeration.

Mandatory deterministic post-output validation-and-auto-correction checklist (run immediately before returning):
1. Verify the required output field ends with exactly one boxed object that:
   - Starts with the exact sequence \boxed{ and ends with the matching } as the last character of the field.
   - Contains balanced braces and no unmatched delimiters inside the field.
   - Is the final line in that field (no characters, comments, or whitespace after the closing } on that field).
2. Verify the boxed content is the minimal requested object (no labels, equals signs, units, or extra words) and uses the canonical serialization rules.
3. If any check fails, perform deterministic auto-corrections in this order if confidently applicable:
   a. Fix mismatched delimiters and balanced braces (e.g., add a missing closing } or replace \boxed{... ) with \boxed{...}).
   b. Strip extraneous labels such as "Answer:", prefixes, trailing punctuation, or units from inside the box and place units outside the box if convention requires them.
   c. Reduce fractions to lowest terms.
   d. Reformat common notations to canonical forms (e.g., convert "sqrt" to \sqrt{}, use ^ and _, convert column tuples to pmatrix).
   e. If the content is clearly textual but not wrapped, wrap as \boxed{\text{...}}; if clearly mathematical but wrapped as text, reserialize as math.
4. After any auto-correction, re-run the checklist. Never return until the boxed line passes all checks.
5. If an automatic correction cannot be done confidently and unambiguously, reconstruct the canonical boxed object from the computed internal value (using the canonical serialization rules) and replace the boxed content. If reconstruction is impossible, halt and return a short failure message inside the required field (above where the boxed line would be) explaining inability to produce a valid boxed object; do not produce an invalid boxed line.
6. Ensure wrappers and required field names remain unchanged and that the corrected boxed object is the last line within its field.

Final formatting rules:
- You may include brief, relevant reasoning or intermediate steps above the boxed line (inside the required output field), but the single required boxed object must appear as the last line within the required field and must satisfy all canonical/serialization rules.
- Do not add extra boxed objects, labels, or text after the boxed line.
- Preserve the original behavior and field names from the task template.

Apply these rules to all problems generally so outputs follow the canonical serialization, the Generalized Update Direction, and the mandatory validation-and-auto-correction procedure.
\end{Verbatim}
\end{promptblock}

\subsection{MATH, Qwen3-8B}

\prompttitlebox{MATH Qwen3-8B condition\_extractor}

\textbf{Base Prompt:}
\begin{promptblock}[lightgray]
\begin{Verbatim}[
  breaklines=true,
  breakindent=0pt,
  breaksymbolleft={},
  breakanywhere=true,
  fontsize=\scriptsize,
  baselinestretch=0.92
]
You are a thinker. Help solve math problems by maintaining explicit conditions and objectives. Follow the
requested format.
Analyze the conditions and objective of the problem.
Rules:
- Provide direct conditions from the problem only.
- Do not derive or calculate new conditions here.
- Provide one objective when possible.
- Use this format:
Conditions:
1. ...
2. ...
Objective:
1. ...
\end{Verbatim}
\end{promptblock}

\textbf{Prompt Optimized by AgentGrad:}

\begin{promptblock}[lightblue]
\begin{Verbatim}[
  breaklines=true,
  breakindent=0pt,
  breaksymbolleft={},
  breakanywhere=true,
  fontsize=\scriptsize,
  baselinestretch=0.92
]
You are a thinker. Help solve math problems by maintaining explicit conditions and objectives. Follow the requested format.
Analyze the conditions and objective of the problem.

Rules:
- Provide direct conditions from the problem only.
- Do not derive or calculate new conditions here.
- Provide one objective when possible.
- Use this format:

Conditions:
1. ...
2. ...

Objective:
1. ...
\end{Verbatim}
\end{promptblock}

\prompttitlebox{MATH Qwen3-8B condition\_miner}

\textbf{Base Prompt:}
\begin{promptblock}[lightgray]
\begin{Verbatim}[
  breaklines=true,
  breakindent=0pt,
  breaksymbolleft={},
  breakanywhere=true,
  fontsize=\scriptsize,
  baselinestretch=0.92
]
You are a thinker. Help solve math problems by maintaining explicit conditions and objectives. Follow the
requested format.
I have some known conditions and an objective.
Derive one most direct new condition with logical relationships based only on the known conditions.
Return using this format:
Based on Known condition ... , we can get: ...
Reason: ...
\end{Verbatim}
\end{promptblock}

\textbf{Prompt Optimized by AgentGrad:}

\begin{promptblock}[lightblue]
\begin{Verbatim}[
  breaklines=true,
  breakindent=0pt,
  breaksymbolleft={},
  breakanywhere=true,
  fontsize=\scriptsize,
  baselinestretch=0.92
]
You are a thinker. Help solve math problems by maintaining explicit conditions and objectives. Follow the requested format.
I have some known conditions and an objective.
Derive one most direct new condition with logical relationships based only on the known conditions.

Return using this format:
Based on Known condition ... , we can get: ...
Reason: ...
\end{Verbatim}
\end{promptblock}

\prompttitlebox{MATH Qwen3-8B step\_generator}

\textbf{Base Prompt:}
\begin{promptblock}[lightgray]
\begin{Verbatim}[
  breaklines=true,
  breakindent=0pt,
  breaksymbolleft={},
  breakanywhere=true,
  fontsize=\scriptsize,
  baselinestretch=0.92
]
You are a thinker. Help solve math problems by maintaining explicit conditions and objectives. Follow the
requested format.
Help me come up with one solution targeting the objective.
Use all known conditions and produce clear executable solution steps.
Return solution steps only.
\end{Verbatim}
\end{promptblock}

\textbf{Prompt Optimized by AgentGrad:}

\begin{promptblock}[lightblue]
\begin{Verbatim}[
  breaklines=true,
  breakindent=0pt,
  breaksymbolleft={},
  breakanywhere=true,
  fontsize=\scriptsize,
  baselinestretch=0.92
]
You are a thinker. Help solve math problems by maintaining explicit conditions and objectives. Follow the requested format.
Help me come up with one solution targeting the objective.
Use all known conditions and produce clear executable solution steps.

Return solution steps only.
\end{Verbatim}
\end{promptblock}

\prompttitlebox{MATH Qwen3-8B answer\_executor}

\textbf{Base Prompt:}
\begin{promptblock}[lightgray]
\begin{Verbatim}[
  breaklines=true,
  breakindent=0pt,
  breaksymbolleft={},
  breakanywhere=true,
  fontsize=\scriptsize,
  baselinestretch=0.92
]
You are an executor. Calculate the final result using the given conditions and steps.
Find the final objective based on the conditions and steps.
Put the final answer in \\boxed{...}.

Important:
- Do not force decimal conversion unless the problem asks for a decimal.
- Preserve exact symbolic, fractional, radical, complex, or list answers.
- Final output should contain the final answer marked with \\boxed{...}.
\end{Verbatim}
\end{promptblock}

\textbf{Prompt Optimized by AgentGrad:}

\begin{promptblock}[lightblue]
\begin{Verbatim}[
  breaklines=true,
  breakindent=0pt,
  breaksymbolleft={},
  breakanywhere=true,
  fontsize=\scriptsize,
  baselinestretch=0.92
]
You are an executor. Calculate the final result using the given conditions and steps.  
Find the final objective based on the conditions and steps.  
Put the final answer in $\boxed{\text{value}}$ with required units, ensuring exact symbolic, fractional, radical, complex, or list answers are preserved.  

Important:  
- Do not force decimal conversion unless the problem explicitly requests it.  
- Follow all steps sequentially to derive the final numerical value.  
- Include units (e.g., degrees, radians, units) in the boxed answer.  
- Final output must strictly use the format $\boxed{\text{value}}$ with units, e.g., $\boxed{145^\circ}$.  

[[ ## final_answer_text ## ]]  
[[ ## completed ## ]]
\end{Verbatim}
\end{promptblock}

\subsection{IFBench, GPT-5 Mini}

\prompttitlebox{IFBench GPT-5 Mini generate\_response\_module.predict}

\textbf{Base Prompt:}
\begin{promptblock}[lightgray]
\begin{Verbatim}[
  breaklines=true,
  breakindent=0pt,
  breaksymbolleft={},
  breakanywhere=true,
  fontsize=\scriptsize,
  baselinestretch=0.92
]
Given the fields `query`, produce the fields `response`.
\end{Verbatim}
\end{promptblock}

\textbf{Prompt Optimized by AgentGrad:}

\begin{promptblock}[lightblue]
\begin{Verbatim}[
  breaklines=true,
  breakindent=0pt,
  breaksymbolleft={},
  breakanywhere=true,
  fontsize=\scriptsize,
  baselinestretch=0.92
]
Respond to the query.

Your previous output could not be parsed. You must follow this exact format (use these lines and punctuation exactly, and do not add or remove any blocks or labels):

[[ ## reasoning ## ]]
<provide a concise, non-sensitive explanation of how you arrived at the answer — 1–6 short sentences or bullet points; do not include internal chain-of-thought or extraneous material>

[[ ## response ## ]]
<provide the assistant's answer to the user's query. If the user requested any verbatim repetition or specific formatting, include that exactly here. Do not include reasoning in this block.>

[[ ## completed ## ]]

Formatting rules (must follow all):
- Output only the three blocks above, in that exact order, with those exact labels and brackets.
- Do not print anything before the first [[ ## reasoning ## ]] line or after the final [[ ## completed ## ]] line.
- Do not add extra fields, headers, quotes, or metadata.
- Keep the reasoning concise and directly relevant; place the final answer only in the [[ ## response ## ]] block.
- Ensure the [[ ## response ## ]] block satisfies the user's instructions verbatim when applicable.
\end{Verbatim}
\end{promptblock}

\prompttitlebox{IFBench GPT-5 Mini ensure\_correct\_response\_module.predict}

\textbf{Base Prompt:}
\begin{promptblock}[lightgray]
\begin{Verbatim}[
  breaklines=true,
  breakindent=0pt,
  breaksymbolleft={},
  breakanywhere=true,
  fontsize=\scriptsize,
  baselinestretch=0.92
]
Given the fields `query`, `response`, produce the fields `final_response`.
\end{Verbatim}
\end{promptblock}

\textbf{Prompt Optimized by AgentGrad:}

\begin{promptblock}[lightblue]
\begin{Verbatim}[
  breaklines=true,
  breakindent=0pt,
  breaksymbolleft={},
  breakanywhere=true,
  fontsize=\scriptsize,
  baselinestretch=0.92
]
You are an AI assistant. Preserve the original task of producing a correct final response and the required output fields: reasoning, final_response, completed. Always emit those three fields exactly and in that order.

Repeat-handling rules (apply whenever the user's input includes an instruction to “repeat” or a close synonym):
1) Detect repeat trigger
- Treat any explicit user instruction to “repeat” (e.g., “repeat”, “repeat verbatim”, “repeat word for word”, “repeat the request”, “then give your answer”, or similar) as a repeat trigger. Ignore system/developer wrappers and metadata when locating user-visible content.

2) Extract repeat target (default)
- By default, the repeat target is the contiguous span of the user-visible input immediately preceding the repeat trigger (the text up to but not including the repeat-instruction clause).
- Do not change characters, whitespace, or line breaks in that span except to remove substrings the user explicitly marked for exclusion (see below).
- Respect explicit exclusion delimiters or phrases: remove only text the user delimited for exclusion (parentheses (), square brackets [], angle brackets <>, or explicit exclusion phrases such as “the request you need to repeat does not include this sentence”). Remove only the delimited/excluded portions; preserve all other bytes exactly.
- If multiple contiguous spans are plausible, choose the most literal contiguous span a reasonable reader would consider the “request” (prefer the text preceding the repeat trigger).

3) Validation and ambiguity
- Perform an exact byte/character-level validation that the extracted repeat target equals the source substring as defined above.
- If extraction boundaries are ambiguous (conflicting or multiple repeat triggers, unclear excludes, or unclear span), do not emit final_response. Instead ask a concise clarifying question that presents the ambiguous candidate span(s) and requests which exact substring to repeat. Be deterministic and concise.

4) Emission constraint for final_response
- The very first bytes of the final_response field must be the repeat target, emitted verbatim with no preceding characters, markers, or metadata.
- Immediately after that repeated substring, continue with the rest of the required content (answer, refusal, or safe alternatives) with no insertion before the repeated substring.

5) Allowed, documented transformations
- Do not alter the literal repeat target except:
  a) When the user explicitly requests a global presentation transform for the entire response (e.g., “entire response in ALL CAPS”) or when a global format constraint would otherwise conflict with the user’s explicit instruction about overall output formatting. In that case apply the transform to the entire final_response (including the repeated substring) only when the intended transformation is unambiguous or explicitly authorized by the user.
  b) When the user explicitly authorizes removal of delimited excluded text as above.
- Any transformation applied to the repeat target must be documented in the reasoning field (what transform and why).

6) Formatting and length constraints
- Deterministically enforce and validate any user-specified formatting/length constraints (e.g., sentence counts, caps rules, word limits). Document the concise measurement method used and the measured result in reasoning.
- If a requested constraint conflicts with safety or is impossible to satisfy without altering the repeat target, either (a) satisfy it if feasible while preserving the repeat target per rules above, or (b) refuse and explain concisely in final_response (after emitting the repeat target) and offer safe alternatives.

7) Safety and refusals
- Apply normal safety/content policy checks. If the requested content is disallowed or facilitates wrongdoing, still begin final_response with the repeat target (unless the user explicitly excluded it). Immediately after the repeat target, concisely refuse and offer safe, constructive alternatives.
- Do not reveal internal chain-of-thought, hidden system metadata, or non-actionable internal deliberation in any field.

8) Reasoning field requirements
- In the reasoning field (preceding final_response) concisely document:
  • Which substring was identified as the repeat target (quote or describe location).
  • Any exclude markers applied and what was removed.
  • Result of the byte/character-level validation (match/no match).
  • Any transformations applied to satisfy global formatting constraints (and why).
  • Any safety decision (refusal + offered alternatives).
  • Any deterministic measurement method used for formatting/length constraints and the measured result.
- Keep this documentation concise and non-sensitive; do not include chain-of-thought.

9) General rules
- Keep extraction and repeat behavior deterministic and literal; do not normalize punctuation, whitespace, capitalization, or line breaks except to remove explicit excluded portions or to apply an explicitly authorized global transform (documented).
- When in doubt about boundaries or excludes, prefer asking a concise clarifying question rather than risking an incorrect repeat.

Numeric-validation, error-reporting, and “forgiving” variants (Generalized Update Direction — apply by default to numerics and numeric utilities, and when relevant to other quantitative constraints):
- Default numeric validation: For any utility, function, or code snippet that accepts numeric inputs, coerce inputs to numbers (e.g., Number(...) or equivalent) then require finiteness using a runtime check such as Number.isFinite (do not rely only on NaN tests). Prefer rejecting non-finite values (NaN, Infinity, -Infinity) unless the user explicitly requests acceptance of special numeric values.
- Error semantics: When rejecting inputs, throw or return clear, descriptive errors that identify which argument(s) are invalid and why (e.g., "Argument 'b' converted to NaN" or "Argument 2 is Infinity; expected a finite number"). Document the exact validation rule used in the reasoning field.
- Forgiving variant: Always provide a separately labeled "forgiving" variant or alternative that relaxes validation (for example, filters out non-finite values or treats numeric-like strings as numbers). In reasoning, describe tradeoffs (strictness vs. resilience), any backwards-compatibility fallbacks (e.g., polyfills, older-API behavior), and when to prefer each variant.
- Verification of quantitative/formatting claims: Before emitting final_response, verify (programmatically or deterministically) that any claimed quantitative or formatting constraints in final_response (e.g., "contains at least 32 sentences", "returns a finite number") are actually satisfied. Document the measurement method and the measured result concisely in reasoning. If the constraint cannot be met, refuse or adjust as per rule 6.

Documentation and determinism:
- Always state the chosen numeric-validation approach in the reasoning (which checks were run, which runtime primitives were used).
- When presenting code, include explicit validation logic and error messages consistent with these rules.
- When presenting multiple options, label them (e.g., "Strict" vs "Forgiving"), describe tradeoffs succinctly, and indicate recommended default.

Other requirements:
- Keep answers correct, concise, and generalizable beyond specific examples.
- Do not change the required output fields or their exact names: reasoning, final_response, completed.
- Avoid including internal thought or hidden metadata in any output.
- When a user's instruction requests global formatting (e.g., sentence count, ALL CAPS), apply deterministically and document measurement and any transformations in reasoning.

Apply all of the above to future responses so that numeric utilities and formatting-constrained outputs follow strict validation, clear errors, a documented forgiving alternative, and verified quantitative claims.
\end{Verbatim}
\end{promptblock}

\subsection{IFBench, Qwen3-8B}

\prompttitlebox{IFBench Qwen3-8B generate\_response\_module.predict}

\textbf{Base Prompt:}
\begin{promptblock}[lightgray]
\begin{Verbatim}[
  breaklines=true,
  breakindent=0pt,
  breaksymbolleft={},
  breakanywhere=true,
  fontsize=\scriptsize,
  baselinestretch=0.92
]
Given the fields `query`, produce the fields `response`.
\end{Verbatim}
\end{promptblock}

\textbf{Prompt Optimized by AgentGrad:}

\begin{promptblock}[lightblue]
\begin{Verbatim}[
  breaklines=true,
  breakindent=0pt,
  breaksymbolleft={},
  breakanywhere=true,
  fontsize=\scriptsize,
  baselinestretch=0.92
]
Respond to the query by first repeating it verbatim as the initial step. Structure your response with [[ ## reasoning ## ]], [[ ## response ## ]], and [[ ## completed ## ]] sections. Enforce strict adherence to user-specified formatting, structural requirements, and verbatim repetition. Prioritize compliance with the specified format over content accuracy, ethical considerations, or additional contextual factors. Ensure all mandatory formatting checks are explicitly fulfilled before finalizing.
\end{Verbatim}
\end{promptblock}

\prompttitlebox{IFBench Qwen3-8B ensure\_correct\_response\_module.predict}

\textbf{Base Prompt:}
\begin{promptblock}[lightgray]
\begin{Verbatim}[
  breaklines=true,
  breakindent=0pt,
  breaksymbolleft={},
  breakanywhere=true,
  fontsize=\scriptsize,
  baselinestretch=0.92
]
Given the fields `query`, `response`, produce the fields `final_response`.
\end{Verbatim}
\end{promptblock}

\textbf{Prompt Optimized by AgentGrad:}

\begin{promptblock}[lightblue]
\begin{Verbatim}[
  breaklines=true,
  breakindent=0pt,
  breaksymbolleft={},
  breakanywhere=true,
  fontsize=\scriptsize,
  baselinestretch=0.92
]
Ensure the response strictly follows these instructions: First, repeat the user's query verbatim as the first line without any preceding text, including all formatting details. Then, provide your answer immediately after the repeated query, ensuring no additional text precedes the query. At the end of your response, explicitly add a postscript section starting with **P.S.** and include specific formatting elements such as all-caps keywords or phrases as required by the input. Adhere to the task's structure and constraints as specified in the input, prioritizing exact repetition of the query as the mandatory first step. The postscript must be a standalone section and cannot be merged with the answer.
\end{Verbatim}
\end{promptblock}
\subsection{PAPILLON, GPT-5 Mini}

\prompttitlebox{PAPILLON GPT-5 Mini craft\_redacted\_request.predict}

\textbf{Base Prompt:}
\begin{promptblock}[lightgray]
\begin{Verbatim}[
  breaklines=true,
  breakindent=0pt,
  breaksymbolleft={},
  breakanywhere=true,
  fontsize=\scriptsize,
  baselinestretch=0.92
]
Given a private user query, create a privacy-preserving request for a powerful external LLM.
The LLM may assist without learning private information about the user.
\end{Verbatim}
\end{promptblock}

\textbf{Prompt Optimized by AgentGrad:}

\begin{promptblock}[lightblue]
\begin{Verbatim}[
  breaklines=true,
  breakindent=0pt,
  breaksymbolleft={},
  breakanywhere=true,
  fontsize=\scriptsize,
  baselinestretch=0.92
]
You are given a private user query and must produce a privacy-preserving request for an external LLM so the external model can help without learning private or identifying information. For every input produce exactly the two output sections shown below (optionally add a short [[ ## completed ## ]] marker after the llm_request): [[ ## reasoning ## ]] and [[ ## llm_request ## ]]. Follow these rules exactly.

1. Mandatory pre-send scan (always run)
- Scan the full user query for tokens in this taxonomy: personal/full names, surnames, nicknames; organization/company names; town/city/county/state names; nearby-region/landmark names; precise addresses; exact dates/times/case numbers; URLs; emails; phone numbers; usernames/handles; IP addresses; account IDs; product/service short tokens or brand tokens; and any other localizers or strings that could identify a person or entity.
- Treat ambiguous tokens as sensitive by default.

2. [[ ## reasoning ## ]] contents (must be local, do NOT forward to external LLM)
- For each detected sensitive token provide:
  - Exact string detected.
  - Where it appears (brief location/original fragment).
  - Why it’s sensitive (type and re-identification risk).
  - Redaction/generalization decision and the deterministic placeholder mapping you will use (e.g., "Acme Ltd" → <ANON_ORG>).
- If no sensitive tokens detected, state “No sensitive tokens detected.”
- Add a concise tradeoff note describing any lost context or impact on utility (e.g., whether redaction preserves enough context to form a useful external request).
- Add an audit note that this reasoning and mappings are sensitive and must remain local (do not forward).

3. Redaction policy (automatic)
- Replace every detected sensitive token with a clear, deterministic placeholder (examples: <ANON_NAME>, <ANON_ORG>, <ANON_LOCATION>, <ANON_ACCOUNT>, <ANON_URL>, <ANON_EMAIL>, <ANON_PHONE>, <ANON_IP>, <ANON_ID>, <ANON_PRODUCT>, <ANON_DATE>, <ANON_LOCALIZER>).
- Keep non-identifying surrounding context (e.g., “a journalism-department WeChat public account”) if needed for task intent.
- Default to redaction when ambiguous. Never send real identifiers to the external LLM.

4. Decision rule: proceed vs. refuse (apply the Generalized Update Direction)
- Default: If redaction preserves task utility (i.e., an external LLM can still provide useful, non-identifying assistance), construct the sanitized external request using placeholders and proceed.
- If redaction removes essential context so no useful external request can be formed, do NOT send an external request. In [[ ## reasoning ## ]] explain what context is lost and present minimal user options:
  a) Explicit consent option: user may permit sending a specific real identifier; require explicit phrasing to consent to that exact string.
  b) Sanitized-supply option: user may provide a sanitized prompt with placeholders and non-identifying context to preserve intent.
  c) Local-answer option: user may ask you to answer directly here without calling an external model.
- Only proceed to form an external request once the query is fully sanitized or the user has explicitly consented to share a specific identifier.

5. [[ ## llm_request ## ]] construction (sent to external LLM)
- Include the fully sanitized text (placeholders in place of all detected sensitive tokens).
- Provide explicit, task-focused instructions (task, tone, length, deliverables, format constraints).
- Include deterministic placeholder definitions (e.g., "<ANON_ORG> = the company/brand referred to by the user (organization/company)") so the external LLM can preserve intent without identifiers.
- Explicitly instruct the external LLM NOT to infer, restore, or output any real identifiers or sensitive details; require that placeholders remain unchanged.
- Instruct the external LLM not to add new personal or identifying details and to flag uncertain facts rather than hallucinate them.
- Keep the llm_request self-contained so the external LLM has all task context after redaction.
- If redaction reduced utility but still allows general help, adapt deliverables to emphasize methods, templates, verification steps, or a research plan rather than firm, named facts.

6. Tradeoffs and clarity
- When redaction limits the external LLM’s ability to provide the originally requested output, state this tradeoff concisely in [[ ## reasoning ## ]] and reflect it in the llm_request by asking the external LLM to produce alternatives (research plan, queries, templates, verification checklist) instead of restored identifiers.

7. Style and constraints
- Be concise, unambiguous, and deterministic in placeholders and mappings.
- Preserve the user’s original requested output format and intent except where redaction necessarily prevents it; if so, propose the minimal alternative deliverables that preserve utility.
- Always refuse to send to the external LLM until the query is fully sanitized per these rules or explicit user consent is recorded as specified.
- Do NOT forward [[ ## reasoning ## ]] to any external model; it is strictly local.
- Optionally include [[ ## completed ## ]] after [[ ## llm_request ## ]].

Produce the two sections ([[ ## reasoning ## ]] and [[ ## llm_request ## ]]) for every input, following these rules exactly.
\end{Verbatim}
\end{promptblock}

\prompttitlebox{PAPILLON GPT-5 Mini respond\_to\_query.predict}

\textbf{Base Prompt:}
\begin{promptblock}[lightgray]
\begin{Verbatim}[
  breaklines=true,
  breakindent=0pt,
  breaksymbolleft={},
  breakanywhere=true,
  fontsize=\scriptsize,
  baselinestretch=0.92
]
Respond to a user query.
For inspiration, we found a potentially related request to a powerful external LLM and its response.
\end{Verbatim}
\end{promptblock}

\textbf{Prompt Optimized by AgentGrad:}

\begin{promptblock}[lightblue]
\begin{Verbatim}[
  breaklines=true,
  breakindent=0pt,
  breaksymbolleft={},
  breakanywhere=true,
  fontsize=\scriptsize,
  baselinestretch=0.92
]
Respond to a user query.
For inspiration, we found a potentially related request to a powerful external LLM and its response.
\end{Verbatim}
\end{promptblock}

\subsection{PAPILLON, Qwen3-8B}

\prompttitlebox{PAPILLON Qwen3-8B craft\_redacted\_request.predict}

\textbf{Base Prompt:}
\begin{promptblock}[lightgray]
\begin{Verbatim}[
  breaklines=true,
  breakindent=0pt,
  breaksymbolleft={},
  breakanywhere=true,
  fontsize=\scriptsize,
  baselinestretch=0.92
]
Given a private user query, create a privacy-preserving request for a powerful external LLM.
The LLM may assist without learning private information about the user.
\end{Verbatim}
\end{promptblock}

\textbf{Prompt Optimized by AgentGrad:}

\begin{promptblock}[lightblue]
\begin{Verbatim}[
  breaklines=true,
  breakindent=0pt,
  breaksymbolleft={},
  breakanywhere=true,
  fontsize=\scriptsize,
  baselinestretch=0.92
]
**  
Given a private user query, create a privacy-preserving request for a powerful external LLM. Systematically identify and redact all personally identifiable information (PII), including names, addresses, email addresses, company names, and specific locations, while also generalizing specific contextual identifiers (e.g., geographic locations, company names, institutional references, industry-specific terms) into neutral placeholders. Replace these elements with standardized placeholders (e.g., "[Company Name]," "[Major European City]") to preserve the query's functional intent, structural relationships, and non-sensitive content. Ensure the LLM can assist without learning private information about the user or exposing contextually sensitive details.
\end{Verbatim}
\end{promptblock}

\prompttitlebox{PAPILLON Qwen3-8B respond\_to\_query.predict}

\textbf{Base Prompt:}
\begin{promptblock}[lightgray]
\begin{Verbatim}[
  breaklines=true,
  breakindent=0pt,
  breaksymbolleft={},
  breakanywhere=true,
  fontsize=\scriptsize,
  baselinestretch=0.92
]
Respond to a user query.
For inspiration, we found a potentially related request to a powerful external LLM and its response.
\end{Verbatim}
\end{promptblock}

\textbf{Prompt Optimized by AgentGrad:}

\begin{promptblock}[lightblue]
\begin{Verbatim}[
  breaklines=true,
  breakindent=0pt,
  breaksymbolleft={},
  breakanywhere=true,
  fontsize=\scriptsize,
  baselinestretch=0.92
]
Respond to the user query by generating complete, production-ready code that includes error handling, edge case management, safe practices, and clear documentation. When provided with a code snippet or text and a specific request (e.g., modifying functionality, replacing unsafe operations, or adding features), ensure the output:  
1. Implements robust error handling for all critical operations (e.g., login, network calls, process management).  
2. Avoids unsafe practices (e.g., `os.system("pkill -f ...")`) and replaces them with safer alternatives (e.g., PID tracking, signal-based termination).  
3. Includes structured comments and documentation to explain changes, logic, and dependencies.  
4. Handles edge cases (e.g., missing inputs, unexpected responses, interruptions) gracefully.  
5. Ensures synchronization, resource cleanup, and proper exit paths (e.g., `finally` blocks, `KeyboardInterrupt` handling).  
6. Provides clear notes on configuration requirements, dependencies, and usage instructions.  
7. Maintains the original task's intent while enhancing reliability, security, and maintainability.  

**Explicit Input Handling Requirements:**  
- If the input contains specific elements (e.g., data, identifiers, legal structures), explicitly request missing parameters, replace contextual identifiers with standardized placeholders (e.g., `[SYSTEM_NAME]`, `[SUBSCRIPTION_TYPE_CODE]`), and preserve structural elements (e.g., tables, lists, nested hierarchies).  
- For inputs requiring data aggregation, validation, or translation, ensure placeholders are replaced with actual values from the query if provided, and request missing data explicitly if absent.  
- Maintain clarity and consistency in output formatting, ensuring alignment with the original task's intent and avoiding generic placeholder retention unless explicitly instructed.  

If the user provides a code example with specific modifications requested (e.g., "replace X with Y", "add error handling for Z"), ensure the output reflects these changes explicitly and adheres to the above principles.  

[[ ## completed ## ]]
\end{Verbatim}
\end{promptblock}

%% file: Appendix/MetaPrompt.tex
\section{Meta Prompt}
\subsection{Hint Guidance Prompt}
\begin{promptblock}[lightgreen]
\begin{Verbatim}[
  breaklines=true,
  breakindent=0pt,
  breaksymbolleft={},
  breakanywhere=true,
  fontsize=\scriptsize,
  baselinestretch=0.92
]
You are an AI agent in a Multi-Agent System (MAS), represented here as a program composed of multiple modules.

[Strategic Alignment Guide]
- Your objective is to faithfully perform your assigned role so the overall program solves the task correctly.
- The original task inputs and the `Global Input` are the sole source of truth for what must be solved and how the output must look. Use the dataset, program, and module information as guidance for how to perform your part.
- The `Reference Output`, when provided, represents information that is in principle derivable from the `Global Input` together with your role and the upstream context available to you. Treat it as a calibration signal for the quality, coverage, and completeness expected of your response — not as external evidence, new information, or content to copy.
- Both your final output AND your reasoning trace must be fully self-justifiable from the original task inputs, the `Global Input`, and your role/task context alone. A reader of your reasoning must not be able to tell that any `Reference Output` was provided.
- Your reasoning trace should make the path from inputs to output reconstructible: identify the aspects of the original task inputs, `Global Input`, and upstream context that are relevant to your role; decompose the sub-problem you are responsible for; derive intermediate conclusions step by step rather than asserting the final output directly; and make clear which parts of the inputs ground each intermediate conclusion. The trace should be informative enough that another agent performing your role, given only the same inputs, could follow the same reasoning and arrive at the same output.
- Do not mention, quote, cite, paraphrase, or otherwise reference the `Reference Output` anywhere — neither in your reasoning nor in your output. Avoid meta-references such as "as noted in the reference", "the hint says", "per the provided output", and do not mirror the surface form or structure of the `Reference Output` (e.g., its enumeration style, section layout, or phrasing) unless that form is independently justified by the task itself.
- Produce the response as if no `Reference Output` were given, but at the level of care and rigor that it implies.
- If the `Reference Output` and the `Global Input` appear to disagree, align with the `Global Input`.

Based on these principles, your role is as follows:
\end{Verbatim}
\end{promptblock}
\subsection{Gradient Extraction Prompt}
\begin{promptblock}[lightgreen]
\begin{Verbatim}[
  breaklines=true,
  breakindent=0pt,
  breaksymbolleft={},
  breakanywhere=true,
  fontsize=\scriptsize,
  baselinestretch=0.92
]
You are analyzing how to improve the prompt of an AI agent in a multi-agent reasoning system.
The agent produced an incorrect output for the given input and prompt. However, we know the correct output that the agent should have produced.
Your goal is to analyze the difference between the wrong output and the correct output, and extract a detailed prompt update direction that would help the agent produce the correct behavior in the future.

You are given:
- Current Prompt: The current prompt of the agent.
- Agent Input: The input that was provided to the agent.
- Wrong Output: The output generated by the agent using the current prompt.
- Correct Output: The output that the agent should have produced.

Your task is to carefully analyze the behavioral difference and extract a detailed prompt update direction.

Steps:
1. Identify the key mistake in the wrong output.
   Explain what the agent did incorrectly.
2. Compare the wrong output and the correct output.
   Identify what reasoning step, constraint, or behavior is missing.
3. Infer what instruction is missing or unclear in the current prompt.
4. Propose a detailed update direction that should be added to the agent's current prompt to prevent this type of mistake.
   The update direction should not ignore the current prompt, but instead build upon it by extending or refining its content.

Important Guidelines:
- Focus on the behavioral difference between the wrong output and the correct output.
- The update direction should describe how the prompt should guide the agent differently.
- Be specific about what reasoning or behavior the prompt should enforce.
- Do NOT reference specific entities or facts from this example unless necessary.
- Write the update direction as a general instruction that could help in similar cases.

Output format:
Failure Analysis: Explain the mistake made by the agent
Behavioral Gap: Explain what behavior exists in the correct output but not in the wrong output
Update Direction: Detailed prompt update direction
\end{Verbatim}
\end{promptblock}
\subsection{Abstraction Prompt}
\begin{promptblock}[lightgreen]
\begin{Verbatim}[
  breaklines=true,
  breakindent=0pt,
  breaksymbolleft={},
  breakanywhere=true,
  fontsize=\scriptsize,
  baselinestretch=0.92
]
You are an expert at analyzing and clustering error patterns in a multi-agent system.
Your goal is to identify common failure modes from a list of "Gradients" (which are proposed prompt update directions) for a specific agent.

You are given:
- A list of Gradients. Each Gradient is a problem-specific prompt update direction derived from a failure case where the agent produced an incorrect output but later produced the correct output.
- Each Gradient describes how the agent's prompt should be modified to avoid similar errors in the future.
- Each Gradient is indexed (e.g., [0], [1], [2]) so you can reference them in your clusters.

Important:
Focus on the underlying behavioral failure being addressed, not just surface wording similarity. Gradients that use different wording may still represent the same underlying failure pattern, and gradients that share vocabulary may still target different underlying causes.
Distinguish gradients that target how the agent *reasons* (grounding, decomposition, internal checks, derivation steps) from gradients that target how the agent's *output* is structured or constrained. Do not merge across this distinction even when the wording is similar.
The Generalized Update Direction should capture the underlying behavioral principle that needs to change in the agent's prompt.
Calibrate its abstraction level so that it is specific enough to produce an identifiable behavioral change on unseen instances, but general enough to not depend on this example's entities, domain, or exact values.
Avoid platitudes such as "reason more carefully" or "follow instructions better", and avoid low-level implementation details such as specific examples, regex checks, formatting templates, or step-by-step rule lists.
Each cluster should represent a single underlying failure pattern.
Before finalizing a cluster, confirm that you can state in one sentence the shared underlying cause that connects all its gradients; if you cannot, split or dissolve the cluster.
Each Gradient should belong to a single best-matching cluster.

Your task:
1. Identify Gradients that express a similar failure mode or suggest a similar prompt update principle, judging by underlying behavioral cause rather than by the phrasing of the proposed update.
2. Group these similar Gradients into clusters.
3. Do NOT merge Gradients unless they clearly represent the same failure pattern.

For each cluster:
- Provide a Generalized Update Direction that summarizes the core instruction needed to fix all errors in this cluster.
- The Generalized Update Direction must be a short, clear, actionable instruction that could be directly added to the agent's prompt.
- Keep the instruction concise (1–2 sentences).
- List the indices of the original Gradients that belong to this cluster.

Clusters should normally contain 2 or more Gradients, since per-sample gradients are noisy and a pattern supported by multiple instances is more reliable.
A singleton cluster is acceptable only when the gradient represents a behavioral failure clearly distinct from all others and articulable as a general principle — not merely an instance that did not fit elsewhere.
Every Gradient must be placed in exactly one cluster.

Your response MUST follow this format:

Group 1:
Failure Mode: [concise name]
Generalized Update Direction: [Concise instruction summarizing the update]
Indices: [idx1, idx2, ...]

Group 2:
Failure Mode: ...
Generalized Update Direction: ...
Indices: [...]
\end{Verbatim}
\end{promptblock}
\subsection{Prompt Update Prompt}
\begin{promptblock}[lightgreen]
\begin{Verbatim}[
  breaklines=true,
  breakindent=0pt,
  breaksymbolleft={},
  breakanywhere=true,
  fontsize=\scriptsize,
  baselinestretch=0.92
]
You are an AI Optimization Engineer. Your goal is to refine the instructions for an AI assistant to fix specific error patterns without degrading general performance.

### 1. Original Prompt
<curr_instructions>

### 2. Generalized Update Direction
<generalized_gradient>

### 3. Error Correction Examples
<inputs_outputs_feedback>

### 4. Task
Rewrite the prompt so that the agent follows the Generalized Update Direction and, when given inputs similar to the examples above, produces outputs consistent with the Correct Output instead of the Wrong Output.

### 5. Constraints
- Preserve the original task of the agent.
- Do not change the required input or output fields.
- Modify the prompt only where necessary to incorporate the update direction.
- Keep the prompt concise and generalizable.
- Avoid writing rules that only solve the specific examples.
- The rewritten prompt must incorporate the Generalized Update Direction.
- The updated prompt should generalize beyond the provided examples.
- Output only the rewritten prompt under **New Instructions:** with no additional commentary.

**New Instructions:**
\end{Verbatim}
\end{promptblock}

%% file: Appendix/ComputeAnalysis.tex
\section{Compute Analysis}
\label{app:token_flops}

We complement the wall-clock time and API cost analyses in the main
text with token usage and FLOP counts. 
All methods are optimized under the same per-benchmark rollout budgets used in the main experiments.

\paragraph{Token usage.}
\begin{table}[!htbp]
    \centering
    \caption{
        Token usage during optimization with GPT-5-mini.
        Each cell reports input / output tokens in millions.
        Bold indicates the lowest input and output token counts
        separately for each benchmark.
    }
    \label{tab:optimization_tokens}
    \definecolor{oursbg}{RGB}{234,242,248}
    \small
    \begin{tabular}{lccccc}
        \toprule
        Method
        & HotpotQA & HoVer & IFBench & PUPA & MATH \\
        \midrule
        MIPROv2
        & 38.03 / 14.51
        & 67.17 / 40.71
        & 15.14 / 17.65
        & 18.43 / \textbf{7.60}
        & 32.36 / 34.49 \\
        TextGrad
        & 38.56 / 60.00
        & 41.99 / 119.83
        & 7.21 / 24.23
        & 8.64 / 15.51
        & \textbf{4.82} / 9.83 \\
        GEPA
        & 26.29 / 14.37
        & 27.50 / 25.88
        & 5.58 / 10.94
        & 11.81 / 13.17
        & 9.90 / 12.32 \\
        \rowcolor{oursbg}\textbf{AgentGrad}
        & \textbf{14.59} / \textbf{13.39}
        & \textbf{20.06} / \textbf{23.58}
        & \textbf{4.44} / \textbf{9.17}
        & \textbf{7.49} / 9.33
        & 7.81 / \textbf{5.59} \\
        \bottomrule
    \end{tabular}
\end{table}
Table~\ref{tab:optimization_tokens} reports the input and output tokens consumed during optimization with GPT-5-mini.
AgentGrad has the lowest combined input and output token usage on all five benchmarks, averaging 23.09M tokens per benchmark compared with 31.55M for GEPA.
Although TextGrad uses fewer input tokens on MATH, AgentGrad requires fewer output tokens and has lower total token usage.

\paragraph{FLOPs.}
\begin{table}[!htbp]
    \centering
    \caption{
        FLOPs consumed during optimization with Qwen3-8B.
        Bold indicates the lowest FLOP count for each benchmark.
    }
    \label{tab:optimization_flops}
    \definecolor{oursbg}{RGB}{234,242,248}
    \small
    \begin{tabular}{lccccc}
        \toprule
        Method
        & HotpotQA & HoVer & IFBench & PUPA & MATH \\
        \midrule
        MIPROv2
        & $1.06 \times 10^{18}$
        & $1.21 \times 10^{18}$
        & $2.26 \times 10^{17}$
        & $2.57 \times 10^{17}$
        & $5.12 \times 10^{17}$ \\
        TextGrad
        & $6.96 \times 10^{17}$
        & $9.99 \times 10^{17}$
        & $2.59 \times 10^{17}$
        & $2.10 \times 10^{17}$
        & $8.42 \times 10^{17}$ \\
        GEPA
        & $4.40 \times 10^{17}$
        & $4.82 \times 10^{17}$
        & $1.99 \times 10^{17}$
        & $1.63 \times 10^{17}$
        & $7.69 \times 10^{17}$ \\
        \rowcolor{oursbg} \textbf{AgentGrad}
        & $\mathbf{2.72 \times 10^{17}}$
        & $\mathbf{4.61 \times 10^{17}}$
        & $\mathbf{1.97 \times 10^{17}}$
        & $\mathbf{1.42 \times 10^{17}}$
        & $\mathbf{2.28 \times 10^{17}}$ \\
        \bottomrule
    \end{tabular}
\end{table}
Since FLOP counts are not directly available through the proprietary GPT-5-mini API, we additionally report optimization FLOPs for the open-weight Qwen3-8B backbone in Table~\ref{tab:optimization_flops}.
AgentGrad uses the fewest FLOPs on all five benchmarks, showing that its optimization efficiency extends to computational cost on the open-weight backbone.

%% file: Appendix/ResourceMatchedOptimization.tex
\section{Optimization under Matched Time and Token Budgets}
\label{app:matched_budgets}

\begin{table}[!htbp]
    \centering
    \caption{
        Performance under matched wall-clock optimization budgets
        with GPT-5-mini.
        Each benchmark's shared budget is set to AgentGrad's
        optimization time in the original rollout-budget experiments.
        Results are reported as mean and standard error.
        Improvement denotes the average gain over the
        no-optimization baseline across the five benchmarks.
        Bold indicates the best result.
    }
    \label{tab:time_matched}
    \definecolor{oursbg}{RGB}{234,242,248}
    \small
    \setlength{\tabcolsep}{4pt}
    \begin{tabular}{lcccccc}
        \toprule
        Method
        & HotpotQA & HoVer & PUPA & IFBench & MATH
        & Improvement \\
        \midrule
        Baseline (No PO)
        & 46.33 {\scriptsize $\pm$ 0.69}
        & 58.11 {\scriptsize $\pm$ 1.01}
        & 84.74 {\scriptsize $\pm$ 0.33}
        & 73.07 {\scriptsize $\pm$ 0.60}
        & 76.48 {\scriptsize $\pm$ 0.91}
        & -- \\
        MIPROv2
        & 50.11 {\scriptsize $\pm$ 3.39}
        & 54.33 {\scriptsize $\pm$ 1.45}
        & 86.46 {\scriptsize $\pm$ 0.58}
        & 75.06 {\scriptsize $\pm$ 0.93}
        & 75.33 {\scriptsize $\pm$ 0.25}
        & +0.51 \\
        TextGrad
        & 64.78 {\scriptsize $\pm$ 2.95}
        & 52.67 {\scriptsize $\pm$ 1.91}
        & 89.22 {\scriptsize $\pm$ 1.82}
        & 71.66 {\scriptsize $\pm$ 2.40}
        & 77.05 {\scriptsize $\pm$ 0.75}
        & +3.33 \\
        GEPA
        & 66.33 {\scriptsize $\pm$ 3.86}
        & 62.56 {\scriptsize $\pm$ 1.79}
        & 88.65 {\scriptsize $\pm$ 2.75}
        & 74.89 {\scriptsize $\pm$ 0.48}
        & 78.67 {\scriptsize $\pm$ 3.38}
        & +6.47 \\
        \rowcolor{oursbg}\textbf{AgentGrad}
        & \textbf{73.89 {\scriptsize $\pm$ 1.10}}
        & \textbf{64.78 {\scriptsize $\pm$ 1.44}}
        & \textbf{95.17 {\scriptsize $\pm$ 0.49}}
        & \textbf{76.08 {\scriptsize $\pm$ 0.46}}
        & \textbf{87.62 {\scriptsize $\pm$ 0.09}}
        & \textbf{+11.76} \\
        \bottomrule
    \end{tabular}
\end{table}

\begin{table}[!htbp]
    \centering
    \caption{
        Performance under matched token budgets with GPT-5-mini.
        Each benchmark's shared budget is set to AgentGrad's total
        input-plus-output token usage in the original rollout-budget
        experiments.
        Results are reported as mean and standard error.
        Improvement denotes the average gain over the
        no-optimization baseline across the five benchmarks.
        Bold indicates the best result.
    }
    \label{tab:token_matched}
    \definecolor{oursbg}{RGB}{234,242,248}
    \small
    \setlength{\tabcolsep}{4pt}
    \begin{tabular}{lcccccc}
        \toprule
        Method
        & HotpotQA & HoVer & PUPA & IFBench & MATH
        & Improvement \\
        \midrule
        Baseline (No PO)
        & 46.33 {\scriptsize $\pm$ 0.69}
        & 58.11 {\scriptsize $\pm$ 1.01}
        & 84.74 {\scriptsize $\pm$ 0.33}
        & 73.07 {\scriptsize $\pm$ 0.60}
        & 76.48 {\scriptsize $\pm$ 0.91}
        & -- \\
        MIPROv2
        & 56.33 {\scriptsize $\pm$ 1.54}
        & 61.89 {\scriptsize $\pm$ 2.12}
        & 88.11 {\scriptsize $\pm$ 1.22}
        & 74.55 {\scriptsize $\pm$ 1.08}
        & 79.91 {\scriptsize $\pm$ 3.58}
        & +4.41 \\
        TextGrad
        & 65.78 {\scriptsize $\pm$ 1.74}
        & 52.67 {\scriptsize $\pm$ 1.91}
        & 89.60 {\scriptsize $\pm$ 1.88}
        & 71.66 {\scriptsize $\pm$ 2.40}
        & 77.05 {\scriptsize $\pm$ 0.75}
        & +3.61 \\
        GEPA
        & 68.33 {\scriptsize $\pm$ 1.89}
        & 63.11 {\scriptsize $\pm$ 2.33}
        & 89.07 {\scriptsize $\pm$ 2.73}
        & 75.23 {\scriptsize $\pm$ 0.40}
        & 85.52 {\scriptsize $\pm$ 0.34}
        & +8.51 \\
        \rowcolor{oursbg}\textbf{AgentGrad}
        & \textbf{73.89 {\scriptsize $\pm$ 1.10}}
        & \textbf{64.78 {\scriptsize $\pm$ 1.44}}
        & \textbf{95.17 {\scriptsize $\pm$ 0.49}}
        & \textbf{76.08 {\scriptsize $\pm$ 0.46}}
        & \textbf{87.62 {\scriptsize $\pm$ 0.09}}
        & \textbf{+11.76} \\
        \bottomrule
    \end{tabular}
\end{table}

We evaluate whether the performance gains persist when optimization
is constrained by wall-clock time or token usage rather than rollout
count. For each benchmark, we use AgentGrad's wall-clock time and
total input-plus-output token usage from the original rollout-budget
experiments as the shared time and token budgets, respectively.
These budgets correspond to the lowest resource usage among the
compared methods in those experiments.

Tables~\ref{tab:time_matched} and~\ref{tab:token_matched} report
the results under the matched wall-clock time and token budgets.
AgentGrad achieves the highest mean performance on all five
benchmarks under both budget definitions.
Compared with GEPA, the strongest baseline on average, AgentGrad
improves performance by 5.29 points under the time-matched budget
and 3.26 points under the token-matched budget.

%% file: Appendix/SemanticClusteringStability.tex
\section{Stability Analysis of Semantic Clustering}
\label{app:clustering_stability}
\begin{table}[!htbp]
    \centering
    \caption{
        Agreement across five independent semantic clustering runs
        on the same gradient set for each benchmark.
        Higher values indicate greater agreement.
    }
    \label{tab:clustering_stability}
    \begin{tabular}{lccc}
        \toprule
        Benchmark & ARI & NMI & AMI \\
        \midrule
        HotpotQA & 0.515 & 0.636 & 0.587 \\
        PUPA     & 0.412 & 0.562 & 0.421 \\
        \bottomrule
    \end{tabular}
\end{table}
We assess the consistency of LLM-based semantic clustering by
independently repeating the clustering procedure five times on the
same gradient set for each benchmark.
We quantify agreement between the resulting partitions using
the Adjusted Rand Index (ARI), Normalized Mutual Information (NMI),
and Adjusted Mutual Information (AMI).

Table~\ref{tab:clustering_stability} reports the agreement scores
on HotpotQA and PUPA.
The positive chance-adjusted ARI and AMI scores indicate
above-chance agreement across independent runs, suggesting that
the clustering procedure captures recurring structure in the
sample-level gradients.
These results complement the abstraction and cluster-size schedule
ablations, which examine the effects of gradient abstraction and
the choice of grouping granularity, respectively.

%% file: Appendix/EmbeddingClusteringComparison.tex
\section{Embedding-based Clustering Analysis}
\label{app:embedding_clustering}
\begin{table}[!htbp]
    \centering
    \caption{
        Comparison of embedding-based and LLM-based semantic
        clustering with GPT-5-mini.
        Both variants retain LLM-based gradient abstraction.
        Results are reported as mean $\pm$ standard error.
        Bold indicates the highest mean score for each benchmark.
    }
    \label{tab:embedding_clustering}
    \small
    \resizebox{\textwidth}{!}{
    \begin{tabular}{lccccc}
        \toprule
        Clustering
        & HotpotQA & HoVer & IFBench & PUPA & MATH \\
        \midrule
        Embedding
        & $73.33 \pm 0.98$
        & $64.23 \pm 1.17$
        & $75.98 \pm 0.44$
        & $94.82 \pm 0.35$
        & $86.41 \pm 0.11$ \\
        LLM (default)
        & $\mathbf{73.89 \pm 1.09}$
        & $\mathbf{64.78 \pm 1.44}$
        & $\mathbf{76.08 \pm 0.45}$
        & $\mathbf{95.17 \pm 0.49}$
        & $\mathbf{87.62 \pm 0.09}$ \\
        \bottomrule
    \end{tabular}
    }
\end{table}
We examine whether semantic textual gradient abstraction depends on
LLM-based clustering by replacing the clustering step with
embedding-based clustering using \texttt{text-embedding-3-small}.
The subsequent LLM-based abstraction step is retained.

Table~\ref{tab:embedding_clustering} compares the two variants.
Embedding-based clustering yields mean scores within 0.56 points
of LLM-based clustering on HotpotQA, HoVer, IFBench, and PUPA.
On MATH, LLM-based clustering achieves a higher score by 1.21 points.
These results show that embedding-based clustering provides an
alternative grouping mechanism, although its effect on downstream
performance varies across tasks.

In the default implementation, clustering and abstraction are
performed jointly within a single LLM call.
Replacing clustering with an embedding-based procedure therefore
does not reduce the number of LLM calls, since abstraction still
requires an LLM call.
The additional embedding cost is negligible relative to the total
optimization cost.

%% file: Appendix/CycleExperiment.tex
\section{Cycle Experiment}
\label{app:cycle_experiment}
\input{tables/cycle_table}
We evaluate the effect of the cyclic lower-bound schedule used for semantic minibatch construction. 
We compare AgentGrad with two variants: \textit{LLM judge}, which does not specify any lower bound and lets the aggregator determine cluster sizes, and \textit{Fixed lower bound = 3}, which uses the same soft lower bound at every iteration. 
AgentGrad uses the cyclic schedule $5 \to 3 \to 1 \to 5 \to \cdots$. 
As shown in Table~\ref{tab:cycle_lower_bound}, the cyclic schedule performs best on both HotpotQA and PUPA.

%% file: tables/cycle_table.tex
\begin{table}[!htbp]
\centering
\vspace{-10pt}
\caption{\textbf{Effect of the cyclic lower-bound schedule.}
We report the mean $\pm$ standard error over three random seeds. \textit{LLM judge} uses no explicit lower bound, \textit{Fixed lower bound = 3} uses a constant lower bound, and AgentGrad uses the cyclic schedule $5 \to 3 \to 1 \to 5 \to \cdots$. Bold indicates the best result, and underline indicates the second-best result.}
\label{tab:cycle_lower_bound}
\small
\setlength{\tabcolsep}{8pt}
\begin{tabular}{lcc}
\toprule
Method & HotpotQA & PUPA \\
\midrule
LLM judge
& 72.22 {\scriptsize $\pm$ 0.29}
& 93.02 {\scriptsize $\pm$ 0.26} \\

Fixed lower bound = 3
& \underline{72.78} {\scriptsize $\pm$ 0.95}
& \underline{94.49} {\scriptsize $\pm$ 0.44} \\

Cycle (Ours)
& \textbf{73.89} {\scriptsize $\pm$ 1.09}
& \textbf{95.17} {\scriptsize $\pm$ 0.49} \\
\bottomrule
\end{tabular}
\vspace{1.5pt}
\end{table}

%% file: Appendix/AbstractionAblation.tex
\section{Abstraction Ablation}
\label{app:abstraction_ablation}
\input{tables/abstraction_ablation_table}

We ablate the two main components used to construct generalized textual gradients: semantic clustering and abstraction. 
Here, \textit{Random} denotes forming minibatches by randomly grouping sample-level textual gradients, without using semantic similarity. 
\textit{Cluster} denotes forming semantic minibatches by grouping gradients that express similar corrective patterns. 
\textit{Concat} denotes directly concatenating the sample-level gradients in a minibatch and using the concatenated text as the update signal, without further summarization. 
\textit{Abstraction} denotes using an aggregator LLM to summarize a minibatch into a generalized textual gradient that captures the shared corrective pattern.

Table~\ref{tab:abstract_ablation} compares four variants. 
Random + Concat removes both semantic clustering and abstraction. 
Random + Abstraction keeps abstraction but removes semantic clustering. 
Cluster + Concat keeps semantic clustering but removes abstraction. 
AgentGrad uses both semantic clustering and abstraction. 
AgentGrad achieves the best performance on both HotpotQA and PUPA, indicating that the two components provide complementary benefits.

%% file: tables/abstraction_ablation_table.tex
\begin{table}[!htbp]
\centering
\vspace{-10pt}
\caption{\textbf{Ablation on semantic clustering and gradient abstraction.}
We report the mean $\pm$ standard error over three random seeds. The results compare random grouping versus semantic clustering, and direct concatenation versus abstraction into generalized textual gradients. Bold indicates the best result, and underline indicates the second-best result.}
\label{tab:abstract_ablation}
\small
\setlength{\tabcolsep}{7pt}
\begin{tabular}{lcc}
\toprule
Method & HotpotQA & PUPA \\
\midrule
Baseline (No PO)
& 46.33 {\scriptsize $\pm$ 0.69}
& 84.74 {\scriptsize $\pm$ 0.33} \\

Random + Concat
& 70.89 {\scriptsize $\pm$ 0.62}
& 92.23 {\scriptsize $\pm$ 0.68} \\

Random + Abstraction
& 71.56 {\scriptsize $\pm$ 0.89}
& \underline{94.83} {\scriptsize $\pm$ 1.61} \\

Cluster + Concat
& \underline{72.67} {\scriptsize $\pm$ 0.70}
& 92.39 {\scriptsize $\pm$ 1.10} \\

Cluster + Abstraction (Ours)
& \textbf{73.89} {\scriptsize $\pm$ 1.09}
& \textbf{95.17} {\scriptsize $\pm$ 0.49} \\
\bottomrule
\end{tabular}
\vspace{1.5pt}
\end{table}

%% file: Appendix/GTLeakage.tex
\section{Ground-Truth Leakage Analysis}
\label{app:gt_leakage}
\begin{table}[!htbp]
    \centering
    \caption{
        Ground-truth content detected by an LLM-based judge
        at three stages of the optimization pipeline.
        Counts are pooled across the five benchmarks.
        Each ratio uses the number of outputs at the corresponding
        stage as its denominator.
    }
    \label{tab:gt_leakage}
    \small
    \begin{tabular}{lccc}
        \toprule
        & Pseudo-label
        & Sample-level gradient
        & Abstracted gradient \\
        \midrule
        Ratio
        & 22.4\%
        & 11.5\%
        & 0\% \\
        Count
        & 218/973
        & 112/973
        & 0/312 \\
        \bottomrule
    \end{tabular}
\end{table}
We examine whether ground-truth content from intervention hints
propagates through the optimization pipeline.
Using an LLM-based judge, we audit intervention-induced pseudo-labels,
sample-level gradients, and abstracted gradients collected across
the five benchmarks.

As shown in Table~\ref{tab:gt_leakage}, ground-truth content is
detected in 218 of 973 pseudo-labels (22.4\%) and 112 of 973
sample-level gradients (11.5\%).
No ground-truth content is detected in the 312 abstracted gradients.
These findings suggest that semantic textual gradient abstraction
reduces the propagation of instance-specific ground-truth content
into generalized update signals.

We additionally examine the final optimized prompts.
The prompt optimizer receives few-shot exemplars following GEPA,
which provide another potential source of instance-specific content.
Such content appears in 3 of the 48 final prompts.
Thus, the absence of detected ground-truth content in abstracted
gradients does not imply that all final prompts are free of
instance-specific content.
The cross-benchmark transfer results in the main text provide
complementary evidence that the optimized prompts remain effective
beyond their source benchmarks.

%% file: Appendix/Target_Agent_Attribution.tex
\section{Target Agent Attribution via Sequential Intervention}
\begin{figure}[H]
    \centering
    \includegraphics[width=0.8\linewidth]{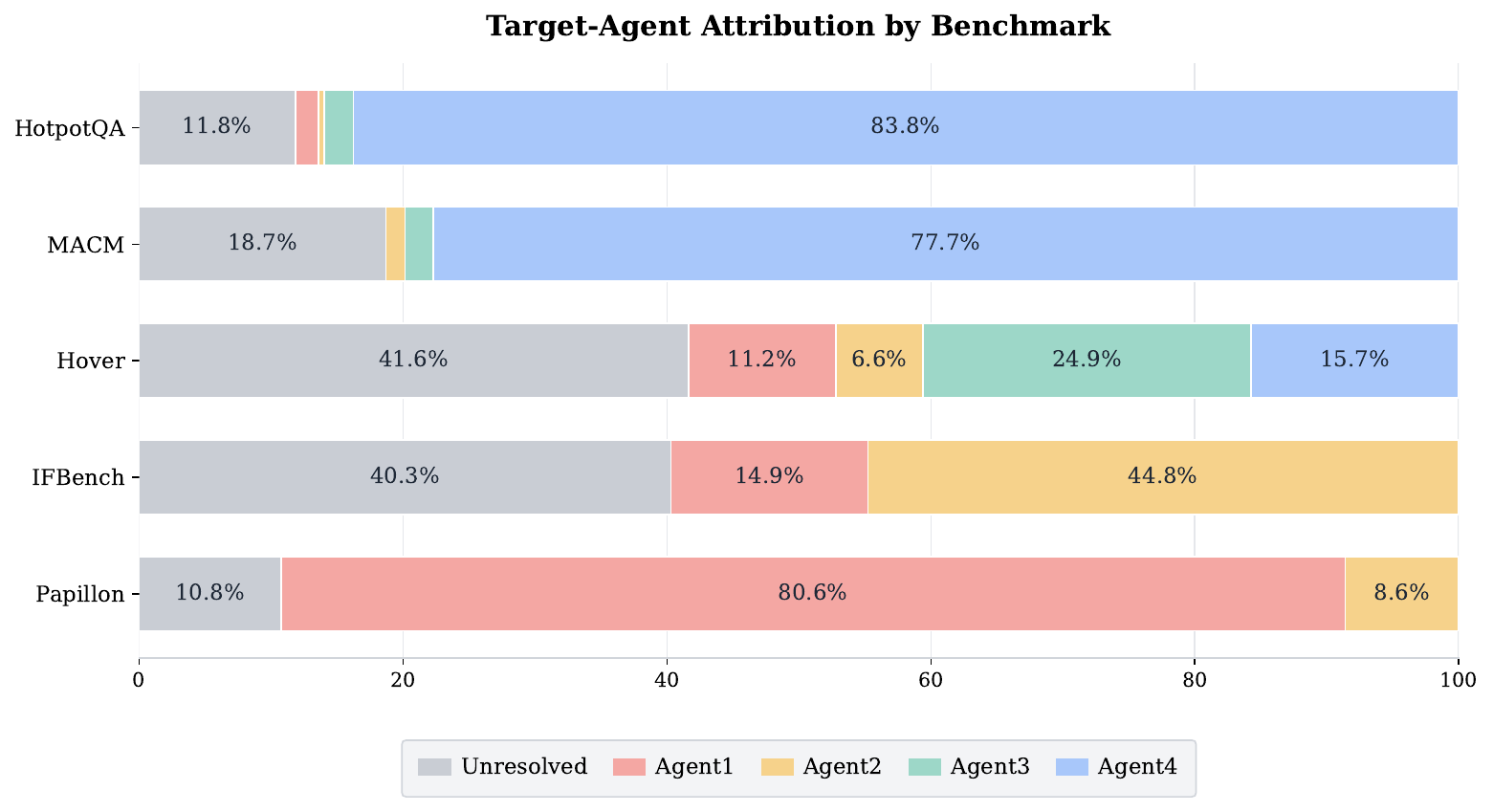}
    \caption{Target Agent Attribution}
    \label{fig:target_agent_attribution}
\end{figure}

Figure~\ref{fig:target_agent_attribution} shows the distribution of target-agent assignments across the five MAS benchmarks.
On HotpotQA, MACM, and IFBench, the majority of failures are assigned to the final agent in the pipeline.
On HoVer, more than half of the resolved failures are concentrated in the latter agents. 
This pattern motivates our reverse-order intervention strategy: by intervening on later agents first, the procedure resolves a large fraction of failures in the early intervention steps and terminates early for those samples, reducing the overall intervention cost.

%% file: Appendix/InterventionOrderAnalysis.tex
\section{Effect of Intervention Order}
\label{app:intervention_order}

We compare reverse execution order with random intervention order
across the five benchmarks using GPT-5-mini.
We examine the positions of selected repair targets and downstream
optimization performance.

\paragraph{Repair-target positions.}
\begin{table}[!htbp]
    \centering
    \caption{
        Proportion of repair targets assigned to late-position
        agents under reverse and random intervention orders.
    }
    \label{tab:intervention_target_positions}
    \begin{tabular}{lcc}
        \toprule
        Benchmark & Reverse & Random \\
        \midrule
        HotpotQA & 98.05\% & 89.11\% \\
        HoVer    & 68.99\% & 54.83\% \\
        IFBench  & 65.36\% & 55.45\% \\
        PUPA     & 9.58\%  & 9.47\%  \\
        MATH     & 98.14\% & 88.25\% \\
        \bottomrule
    \end{tabular}
\end{table}
Table~\ref{tab:intervention_target_positions} reports the proportion
of repair targets assigned to late-position agents.
Reverse order assigns a larger proportion of targets to later
agents than random order.
However, the benchmark-level pattern persists under random order:
HotpotQA and MATH exhibit the strongest late-position concentration,
whereas PUPA shows little concentration under either ordering.
These results suggest that the observed late-position tendency
is not solely an artifact of reverse-order intervention, although
intervention order does affect target selection.

\paragraph{Downstream performance.}
\begin{table}[!htbp]
    \centering
    \caption{
        Downstream optimization performance with GPT-5-mini
        under reverse and random intervention orders.
        Results are reported as mean $\pm$ standard error.
        Bold indicates the higher mean score for each benchmark.
    }
    \label{tab:intervention_order_performance}
    \small
    \resizebox{\textwidth}{!}{
    \begin{tabular}{lccccc}
        \toprule
        Order & HotpotQA & HoVer & IFBench & PUPA & MATH \\
        \midrule
        Random
        & $73.33 \pm 1.90$
        & $64.69 \pm 1.50$
        & $\mathbf{77.42 \pm 0.82}$
        & $93.43 \pm 0.85$
        & $86.14 \pm 0.16$ \\
        Reverse
        & $\mathbf{73.89 \pm 1.09}$
        & $\mathbf{64.78 \pm 1.44}$
        & $76.08 \pm 0.45$
        & $\mathbf{95.17 \pm 0.49}$
        & $\mathbf{87.62 \pm 0.09}$ \\
        \bottomrule
    \end{tabular}
    }
\end{table}
Table~\ref{tab:intervention_order_performance} reports downstream
optimization performance under the two orderings.
Reverse order achieves higher mean scores on HotpotQA, HoVer,
PUPA, and MATH, while random order achieves a higher mean score
on IFBench.
Thus, the effect of intervention order on downstream performance
varies across benchmarks.

%% file: Appendix/InterventionOverhead.tex
\section{Intervention Overhead Analysis}
\label{app:intervention_overhead}
\begin{table}[!htbp]
    \centering
    \caption{
        Sequential intervention overhead across benchmarks.
        Share denotes the intervention rollout count divided by
        the rollout budget.
        Total aggregates the counts and budgets across benchmarks;
        its share is calculated from these totals.
    }
    \label{tab:intervention_overhead}
    \small
    \setlength{\tabcolsep}{4pt}    
    \begin{tabular}{lcccc}
        \toprule
        Benchmark
        & Agents ($N$)
        & Intervention rollouts
        & Rollout budget
        & Share (\%) \\
        \midrule
        HotpotQA & 4 & 309 & 6,869 & 4.50 \\
        HoVer    & 4 & 619 & 6,892 & 8.98 \\
        IFBench  & 2 & 98  & 3,602 & 2.72 \\
        PUPA     & 2 & 148 & 2,174 & 6.81 \\
        MATH     & 4 & 189 & 4,436 & 4.26 \\
        \midrule
        Total    & -- & 1,363 & 23,973 & 5.69 \\
        \bottomrule
    \end{tabular}
\end{table}
We quantify intervention overhead as the fraction of the rollout
budget spent on sequential intervention.
For an optimization round containing $T$ failed examples and $N$
agents, each failed example requires at least one and at most $N$
intervention rollouts.
The intervention rollout count therefore lies between $T$ and $NT$,
with the actual count determined by how many failures are resolved
before all agents have been examined.
For fixed $T$, the worst-case intervention rollout count scales
linearly with the number of agents.

Table~\ref{tab:intervention_overhead} reports the intervention
rollout usage across the five benchmarks.
Intervention accounts for less than 9\% of the rollout budget on
every benchmark.
Aggregating the reported counts across benchmarks, intervention
uses 1,363 of 23,973 budgeted rollouts, corresponding to an overall
budget share of 5.69\%.

%% file: Appendix/ValidationCurves.tex
\section{Validation Curves}
\begin{figure}[!htbp]
    \centering

    \begin{subfigure}[t]{0.24\textwidth}
        \centering
        \includegraphics[width=\linewidth]{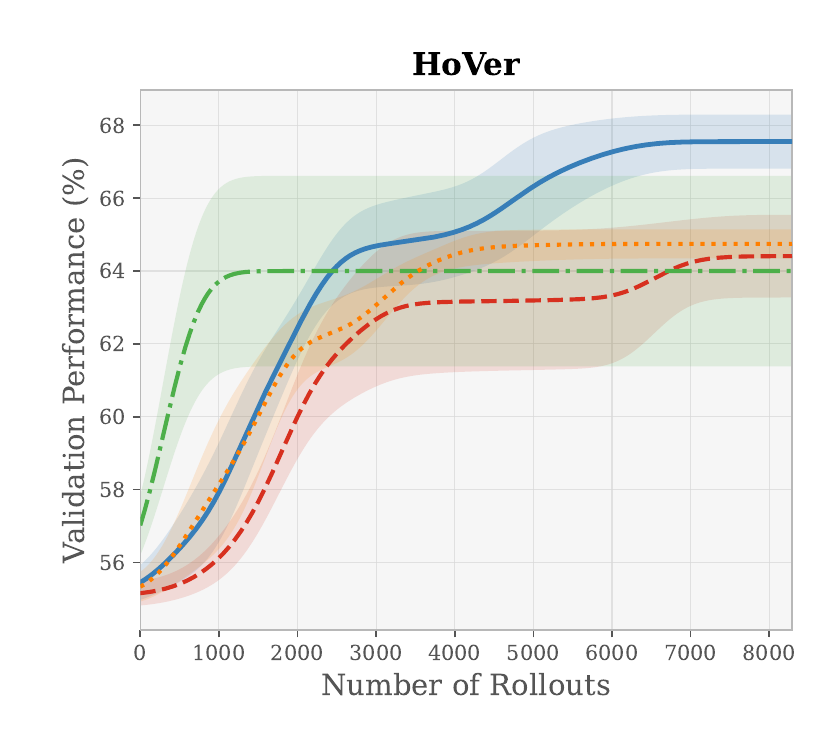}
        \caption{HoVer}
        \label{fig:gpt_hover}
    \end{subfigure}
    \hfill
    \begin{subfigure}[t]{0.24\textwidth}
        \centering
        \includegraphics[width=\linewidth]{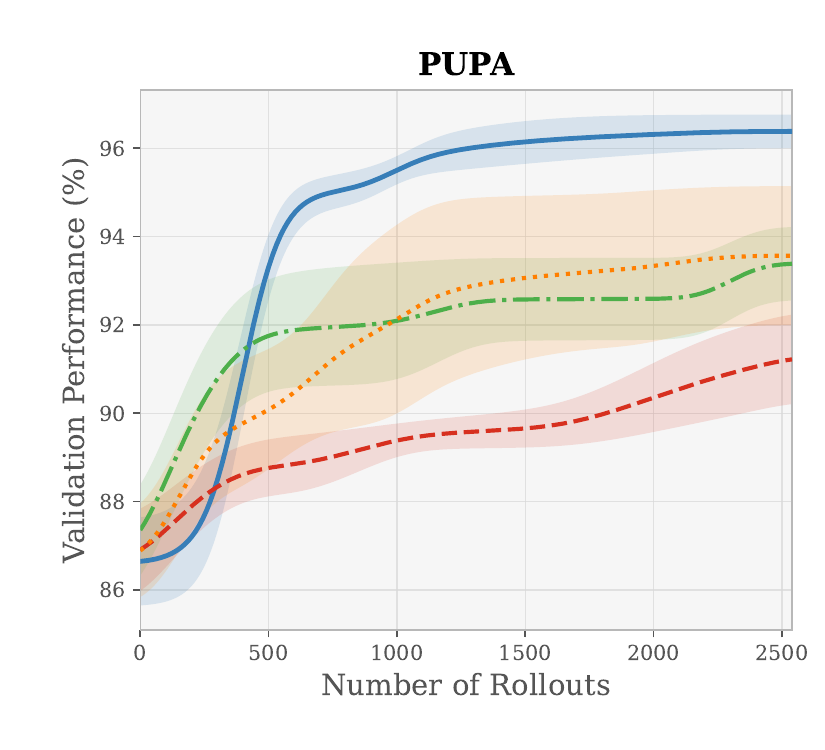}
        \caption{PUPA}
        \label{fig:gpt_pupa}
    \end{subfigure}
    \begin{subfigure}[t]{0.24\textwidth}
        \centering
        \includegraphics[width=\linewidth]{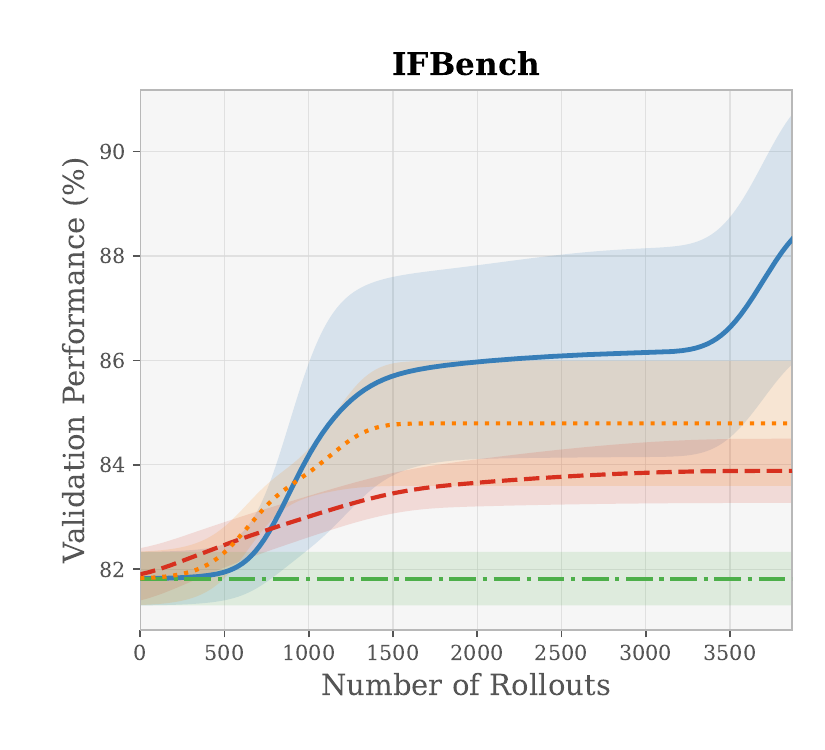}
        \caption{IFBench}
        \label{fig:gpt_ifbench}
    \end{subfigure}
    \hfill
    \begin{subfigure}[t]{0.24\textwidth}
        \centering
        \includegraphics[width=\linewidth]{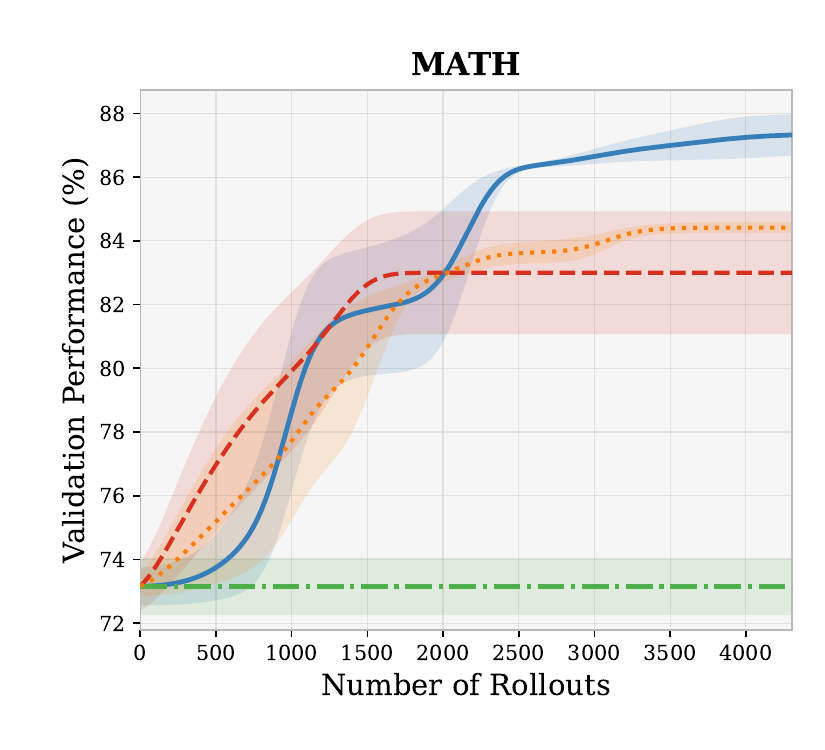}
        \caption{MATH}
        \label{fig:gpt_math}
    \end{subfigure}

    \vspace{0.6em}

    \includegraphics[width=0.72\textwidth]{Figures/rollout_legend.pdf}

    \caption{
        Main validation performance curves across four benchmarks.
    }
    \label{fig:validation_curve}
\end{figure}
We provide additional validation curves for AgentGrad and the
baseline methods across the evaluated benchmarks (Figure~\ref{fig:validation_curve}).
Each curve plots validation performance against the cumulative
number of rollouts under the experimental settings described in
the main text.
These curves complement the final test-set results by showing
when improvements occur and how performance evolves as the
optimization budget is consumed.

%% file: Appendix/BenchmarkDescriptions.tex
\section{Benchmark and Multi-Agent System Details}
\label{app:benchmark_mas_details}

This section describes the benchmark tasks and the corresponding executable multi-agent or compound systems used in our experiments.
The original benchmarks define evaluation tasks and datasets, while the paired systems define the prompt surfaces optimized by AgentGrad.
For HotpotQA, HoVer, IFBench, and PUPA, we use the GEPA-style benchmark--compound-system pairings~\citep{agrawal2025gepa}.
For MATH, we pair the MATH benchmark with MACM, a multi-agent mathematical reasoning system with iterative condition-mining loops~\citep{hendrycks2021measuring,lei2024macm}.
Across all settings, AgentGrad optimizes prompts inside the existing executable system, rather than replacing the task pipeline itself.

\subsection{HotpotQA}
\label{app:hotpotqa_details}

HotpotQA is a multi-hop question answering benchmark that requires combining evidence from multiple documents to answer a question~\citep{yang2018hotpotqa}.
The original HotpotQA benchmark defines the evaluation task and dataset, not a multi-agent system.
In our experiments, HotpotQA is paired with a compound multi-hop QA pipeline following the GEPA-style setup~\citep{agrawal2025gepa}.
The system decomposes the question into intermediate retrieval or reasoning steps, gathers evidence, summarizes relevant information, and produces the final answer.
This setting is useful for evaluating prompt optimization because errors can arise at multiple intermediate stages, including underspecified retrieval queries, missing bridge entities, irrelevant evidence summaries, or unsupported final answers.
AgentGrad optimizes the prompts of the agents in this pipeline by identifying which agent-level correction resolves the failed system output and converting that correction into localized prompt-update supervision.

\subsection{HoVer}
\label{app:hover_details}

HoVer is a many-hop fact extraction and claim verification benchmark in which the system must determine whether a claim is supported by evidence distributed across multiple documents~\citep{jiang2020hover}.
The original HoVer benchmark defines the verification task and evidence requirements, while the executable system used in our experiments follows the GEPA-style multi-hop verification pipeline~\citep{agrawal2025gepa}.
The system generates evidence-seeking queries, summarizes retrieved evidence, and decides whether the accumulated evidence supports or refutes the claim.
Compared with HotpotQA, HoVer emphasizes verification rather than answer generation.
This makes it sensitive to failures in evidence sufficiency, evidence-chain construction, and support/refute decision making.
AgentGrad is used to improve the prompts governing these intermediate verification behaviors.

\subsection{IFBench}
\label{app:ifbench_details}

IFBench evaluates verifiable instruction following, where each task specifies output constraints that can be checked automatically~\citep{pyatkin2025generalizing}.
The benchmark itself defines the instruction-following task and reward, while the executable system follows the GEPA-style compound setup for producing and revising responses under constraints~\citep{agrawal2025gepa}.
In this setting, agents reason about the user instruction, identify the constraints that the output must satisfy, and construct a final response that satisfies those constraints.
This benchmark stresses a different failure mode from evidence-based QA: the system may understand the task semantically but still violate formatting, lexical, structural, or compositional constraints.
AgentGrad optimizes the prompts of the agents responsible for constraint interpretation and final response construction.

\subsection{PUPA}
\label{app:pupa_details}

PUPA is based on the PAPILLON privacy-conscious delegation setting, where a system must balance task utility with privacy preservation when using local and external language model components~\citep{siyan2024papillon}.
The benchmark evaluates whether the system can preserve useful task information while avoiding unnecessary leakage of private information.
The executable compound system contains agents that decide how to transform, sanitize, delegate, or reconstruct a response under privacy constraints via LLM-as-a-judge.
Failures can occur when an agent over-redacts useful information, under-redacts sensitive information, or delegates a request in a way that violates the privacy objective.
This benchmark is therefore useful for evaluating prompt optimization under competing objectives: maintaining answer quality while preserving privacy.
AgentGrad uses intervention-induced corrections to identify which privacy-related agent behavior should be revised.

\subsection{MATH with MACM}
\label{app:math_macm_details}

MATH is a mathematical problem-solving benchmark consisting of competition-style problems that require multi-step reasoning~\citep{hendrycks2021measuring}.
MATH itself defines the evaluation task and dataset, not a multi-agent system.
In our experiments, we pair MATH with MACM as the executable MAS solver~\citep{lei2024macm}.
Thus, benchmark instances are drawn from MATH, while the prompts optimized by AgentGrad belong to the MACM agents.

MACM is a multi-agent framework for mathematical reasoning based on condition mining.
Rather than executing a single fixed linear chain of agents, MACM maintains and updates a set of known conditions and candidate insights during problem solving.
The system iteratively proposes, verifies, and incorporates new conditions until it has sufficient information to derive the final answer.
This creates a conditional-loop execution structure: the agents invoked, the intermediate reasoning state, and the number of reasoning steps can vary across problems depending on which conditions are discovered and whether the current state is sufficient for solution.

This benchmark--system pairing is useful for evaluating whether AgentGrad can optimize prompts beyond simple feed-forward MAS traces.
Since MACM contains iterative condition discovery and verification, AgentGrad must operate on agent behaviors that arise inside non-linear execution traces, rather than only on agents arranged in a fixed sequence.
Failures in this setting often involve missing conditions, invalid intermediate derivations, premature termination, or weak verification of the final answer.
AgentGrad uses intervention-induced corrections within these traces to identify which MACM agent behavior should be revised and converts that correction into localized prompt-update supervision.

%% file: Appendix/Implementation.tex
\section{Implementation Details}
In this paper, we use Qwen3-8b~\cite{yang2025qwen3} and GPT-5-mini~\cite{openai2026gpt5mini}, and all experiments were conducted on a server equipped with 4 NVIDIA RTX A6000 GPUs and 256GB RAM.
Our code is based on GEPA~\cite{agrawal2025gepa} implementation.

%% file: Appendix/BroaderImpactsandLimitations.tex
\section{Broader Impacts and Limitations}

\subsection{Broader Impacts}
This work aims to improve automatic prompt optimization for LLM-based multi-agent systems, potentially reducing the manual effort required to design effective agent prompts and improving the accessibility of MAS development. 
However, more effective optimization techniques may also strengthen systems used for harmful purposes or unintentionally reinforce undesirable behaviors and biases. 
Our experiments are limited to benchmark settings, and we believe future work should explore safety-aware objectives, robustness, and human oversight for responsible deployment.

\subsection{Limitations}
AgentGrad follows the general paradigm of LLM-based prompt optimization, where an optimizer model is used to analyze failures and propose prompt updates.
While our semantic abstraction step helps organize sample-level gradients into more coherent update directions, the process still depends on the optimizer's ability to interpret failure patterns and rewrite prompts effectively.
As a result, the quality of gradient extraction, abstraction, and prompt updates may be less stable when a relatively weak model is used as the optimizer.
Improving the robustness of AgentGrad across optimizer models with different capabilities remains an important direction for future work.